\pdfoutput=1

\documentclass[11pt]{article}

\usepackage[]{acl}

\usepackage{times}
\usepackage{latexsym}

\usepackage[T1]{fontenc}

\usepackage[utf8]{inputenc}

\usepackage{microtype}

\usepackage{booktabs}
\usepackage{graphicx}
\usepackage{enumitem}
\usepackage{xcolor}
\usepackage{multirow}
\usepackage{multicol}
\usepackage{amsmath}
\usepackage{bold-extra}
\usepackage{subcaption}
\usepackage{float}
\usepackage{nicefrac}
\usepackage{color, colortbl}

\DeclareMathOperator*{\argmax}{arg\,max}
\definecolor{linen}{rgb}{0.98, 0.94, 0.9}

\title{Training Dynamics for Curriculum Learning: \\A Study on Monolingual and Cross-lingual NLU}

\author{
Fenia Christopoulou, Gerasimos Lampouras, Ignacio Iacobacci \\
Huawei Noah's Ark Lab, London, UK \\
\small \texttt{\{efstathia.christopoulou, gerasimos.lampouras, ignacio.iacobacci\}@huawei.com} \\
}

\begin{document}
\maketitle

\begin{abstract}
	Curriculum Learning (CL) is a technique of training models via ranking examples in a typically increasing difficulty trend with the aim of accelerating convergence and improving generalisability.
	Current approaches for Natural Language Understanding (NLU) tasks use CL to improve in-distribution data performance often via heuristic-oriented  or task-agnostic difficulties.
	In this work, instead, we employ CL for NLU by taking advantage of training dynamics as difficulty metrics, i.e. statistics that measure the behavior of the model at hand on specific task-data instances during training and propose modifications of existing CL schedulers based on these statistics.
	Differently from existing works, we focus on evaluating models on in-distribution (ID), out-of-distribution (OOD) as well as zero-shot (ZS) cross-lingual transfer datasets.
	We show across several NLU tasks that CL with training dynamics can result in better performance mostly on zero-shot cross-lingual transfer and OOD settings with improvements up by 8.5\% in certain cases.
	Overall, experiments indicate that training dynamics can lead to better performing models with smoother training compared to other difficulty metrics while being 20\% faster on average. In addition, through analysis we shed light on the correlations of task-specific versus task-agnostic metrics\footnote{Code is available at  \url{https://github.com/huawei-noah/noah-research/tree/master/NLP/TD4CL}}.
\end{abstract}

\section{Introduction}

Transformer-based language models \citep[LMs]{vaswani-etal-2017-attention,devlin-etal-2019-bert} have recently achieved great success in a variety of NLP tasks~\citep{wang-etal-2018-glue,wang2019superglue}. However, generalisation to out-of-distribution (OOD) data and zero-shot cross-lingual transfer still remain a challenge~\citep{linzen-2020-accelerate,hu2020xtreme}.
Among existing techniques, improving OOD performance has been addressed by training with adversarial data~\citep{yi2021improved}, while better transfer across languages has been achieved by selecting appropriate languages to transfer from~\citep{lin-etal-2019-choosing,turc2021revisiting}, employing meta-learning~\citep{nooralahzadeh-etal-2020-zero} or data alignment~\citep{fang2020filter}.

Contrastive to such approaches that take advantage of additional training data is Curriculum Learning~\cite[CL]{bengio2009curriculum}, a technique that aims to train models using a specific ordering of the original training examples. This ordering typically follows an increasing difficulty trend where easy examples are fed to the model first, moving towards harder instances. The intuition behind CL stems from human learning, as humans focus on simpler concepts before learning more complex ones, a procedure that is called shaping~\citep{krueger2009flexible}.
Although curricula have been primarily used for Computer Vision~\citep{pmlr-v97-hacohen19a,wu2021when} and Machine Translation~\citep{zhang-etal-2019-curriculum,platanios-etal-2019-competence}, there are only a handful of approaches that incorporate CL into Natural Language Understanding tasks~\citep{sachan-xing-2016-easy,tay-etal-2019-simple,lalor-yu-2020-dynamic,xu-etal-2020-curriculum}.

Typically, CL requires a measure of difficulty for each example in the training set. Existing methods using CL in NLU tasks rely on heuristics such as sentence length, word rarity, depth of the dependency tree~\citep{platanios-etal-2019-competence,tay-etal-2019-simple}, metrics based on item-response theory~\citep{lalor-yu-2020-dynamic} or task-agnostic model metrics such as perplexity~\citep{zhou-etal-2020-uncertainty}.
Such metrics have been employed to either improve in-distribution performance on NLU or Machine Translation. However, their effect is still under-explored on other settings.

In this study instead, we propose to adopt training dynamics~\citep[TD]{swayamdipta-etal-2020-dataset} as difficulty measures for CL and fine-tune models with curricula on downstream tasks.
TD were recently proposed as a set of statistics collected during the course of a model's training to automatically evaluate dataset quality, by identifying annotation artifacts.
These statistics, offer a 3-dimensional view of a model's uncertainty towards each training example classifying them into distinct areas--\textit{easy}, \textit{ambiguous} and \textit{hard} examples for a model to learn.

We test a series of easy-to-hard curricula using TD, namely TD-CL, with existing schedulers as well as novel modifications of those and experiment with other task-specific and task-agnostic metrics.
We show performances and training times on three settings: in-distribution (ID), out-of-distribution (OOD) and zero-shot (ZS) transfer to languages different than English.
To the best of our knowledge, no prior work on NLU considers the impact of CL on all these settings.
To consolidate our findings, we evaluate models on different classification tasks, including Natural Language Inference, Paraphrase Identification, Commonsense Causal Reasoning and Document Classification.

Our findings suggest that TD-CL provides better zero-shot cross-lingual transfer up to 1.2\% over prior work and can gain an average speedup of 20\%, up to 51\% in certain cases.
In ID settings CL has minimal to no impact, while in OOD settings models trained with TD-CL can boost performance up to 8.5\% on a different domain.
Finally, TD provide more stable training compared to another task-specific metric (Cross-Review). On the other hand, heuristics can also offer improvements especially when testing on a completely different domain.

\section{Related Work}

Curriculum Learning was initially mentioned in the work of \citet{elman1993learning} who demonstrated the importance of feeding neural networks with small/easy inputs at the early stages of training. The concept was later formalised by \citet{bengio2009curriculum} where training in an easy-to-hard ordering was shown to result in faster convergence and improved performance.
In general, Curriculum Learning requires a \textit{difficulty metric} (also known as the scoring function) used to rank training instances, and a \textit{scheduler} (known as the pacing function) that decides when and how new examples--of different difficulty--should be introduced to the model.

\noindent \textbf{Example Difficulty} was initially expressed via model loss, in self-paced learning~\citep{kumar2010self-paced,jiang-etal-2015-selfpacedcl}, increasing the contribution of harder training instances over time.
This setting posed a challenge due to the fast-changing pace of the loss during training, thus later approaches used human-intuitive difficulty metrics, such as sentence length or the existence of rare words~\citep{platanios-etal-2019-competence} to pre-compute difficulties of training instances.
However, as such metrics do not express difficulty of the model, model-based metrics have been proposed over the years, such as measuring the loss difference between two checkpoints~\citep{xu-etal-2020-dynamic} or model translation variability~\citep{wang-etal-2019-improving-back,wan-etal-2020-self}.
In our curricula we use training dynamics to measure example difficulty, i.e. metrics that consider difficulty from the perspective of a model towards a certain task.
Example difficulty can be also estimated either in a static (offline) or dynamic (online) manner, where in the latter training instances are evaluated and re-ordered at certain times during training, while in the former the difficulty of each example remains the same throughout. In our experiments we adopt the first setting and consider static example difficulties.

\noindent \textbf{Transfer Teacher CL} is a particular family of such approaches that use an external model (namely the teacher) to measure the difficulty of training examples.
Notable works incorporate a simpler model as the teacher~\citep{zhang2018empirical} or a larger-sized model~\citep{pmlr-v97-hacohen19a}, as well as using similar-sized learners trained on different subsets of the training data.
These methods have considered as example difficulty, either the teacher model perplexity~\citep{zhou-etal-2020-uncertainty}, the norm of a teacher model word embeddings~\citep{liu-etal-2020-norm}, the teacher's performance on a certain task~\citep{xu-etal-2020-curriculum} or simply regard difficulty as a latent variable in a teacher model~\citep{lalor-yu-2020-dynamic}.
In the same vein, we also incorporate Transfer Teacher CL via teacher and student models of the same size and type. However, differently, we take into account the behavior of the teacher \textit{during the course of its training} to measure example difficulty instead of considering its performance at the end of training or analysing internal embeddings.

\noindent Moving on to \textbf{Schedulers}, these can be divided into discrete and continuous. Discrete schedulers, often referred to as \textit{bucketing}, group training instances that share similar difficulties into distinct sets. Different configurations include accumulating buckets over time~\citep{cirik2016visualizing}, sampling a subset of data from each bucket~\citep{xu-etal-2020-curriculum,kocmi-bojar-2017-curriculum} or more sophisticated sampling strategies~\citep{zhang2018empirical}.
In cases where the number of buckets is not obtained in a straightforward manner, methods either heuristically split examples~\citep{zhang2018empirical}, adopt uniform splits~\citep{xu-etal-2020-curriculum} or employ schedulers that are based on a continuous function. A characteristic approach is that of \citet{platanios-etal-2019-competence} where at each training step a monotonically increasing function chooses the amount of training data the model has access to, sorted by increasing difficulty.
As we will describe later on, we experiment with two established schedulers and propose modifications of those based on training dynamics.

Other tasks where CL has been employed include Question Answering~\citep{sachan-xing-2016-easy}, Reading comprehension~\citep{tay-etal-2019-simple} and other general NLU classification tasks~\citep{lalor-yu-2020-dynamic,xu-etal-2020-curriculum}.
Others have developed modified curricula in order to train models for code-switching~\citep{choudhury-etal-2017-curriculum}, anaphora resolution~\citep{stojanovski-fraser-2019-improving}, relation extraction~\citep{huang-du-2019-self}, dialogue~\citep{saito-2018-curriculum,shen-feng-2020-cdl} and self-supervised Neural Machine Translation~\citep{ruiter-etal-2020-self}, while more advanced approaches combine it with Reinforcement Learning in a collaborative teacher-student transfer curriculum~\citep{kumar-etal-2019-reinforcement}.

\section{Methodology}

Let $D = \{(x_i, y_i)\}_{i=1}^N$ be a set of training data instances.
A curriculum is comprised of two main elements: the \textit{difficulty metric}, responsible for associating a training example to a score that represents a notion of difficulty and the \textit{scheduler} that determines the type and number of available instances at each training step $t$.
We experiment with three difficulty metrics derived from training dynamics and four schedulers: two are new contributions and the remaining are referenced from previous work.

\subsection{Difficulty Metrics}
As aforementioned, we use training dynamics~\citep{swayamdipta-etal-2020-dataset}, i.e. statistics originally introduced to analyse dataset quality, as difficulty metrics.
The suitability of such statistics to serve as difficulty measures for CL is encapsulated in three core aspects.
Firstly, training dynamics are straightforward. They can be easily obtained by training a single model on the target dataset and keeping statistics about its predictions on the training set.
Secondly, training dynamics correlate well with model uncertainty and follow a similar trend to human (dis)agreement in terms of data annotation, essentially combining the view of both worlds.
Finally, training dynamics manifest a clear pattern of separating instances into distinct areas--\textit{easy}, \textit{ambiguous} and \textit{hard} examples for a model to learn--something that aligns well with the ideas behind Curriculum Learning.

The difficulty of an example (${x}_i, y_i$) can be determined by a function $f$, where an example $i$ is considered more difficult than example $j$ if $f({x}_i, y_i) > f({x}_j, y_j)$.
We list three difficulty metrics that use statistics during the course of a model's training, as follows:

\setlength{\belowdisplayskip}{2.5pt} \setlength{\belowdisplayshortskip}{2.5pt}
\setlength{\abovedisplayskip}{2.5pt} \setlength{\abovedisplayshortskip}{2.5pt}

\begin{itemize}[leftmargin=-0.5pt, topsep=1pt, itemsep=1pt]
	\item[] \textbf{\textsc{confidence}} \textsc{(conf)} of an example ${x}_i$ is the average probability assigned to the gold label $y_i$ by a model with parameters $\theta$ across a number of epochs $E$. This is a continuous metric with higher values corresponding to easier examples.
	\begin{equation}
		f_\text{\textsc{conf}}(x_i, y_i) = \mu_i = \frac{1}{E} \sum_{e=1}^{E} p_{\theta^{(e)}}(y_i | x_i)
		\label{eq:confidence}
	\end{equation}

	\item[] \textbf{\textsc{correctness}} \textsc{(corr)} is the number of times a model classifies example ${x}_i$ correctly across its training. It takes values between $0$ and $E$. Higher correctness indicates easier examples for a model to learn.
	\begin{multline}
		f_\text{\textsc{corr}}(x_i, y_i) = \sum_{e=1}^E o_i^{(e)},  \\
		o_i^{(e)} =
		\begin{cases}
			1    & \text{if } \argmax p_{\theta^{(e)}}(x_i) = y_i \\
			0,   & \text{otherwise}
		\end{cases}
		\label{eq:correctness}
	\end{multline}

	\item[] \textbf{\textsc{variability}} \textsc{(var)} of an example ${x}_i$ is the standard deviation of the probabilities assigned to the gold label $y_i$ across $E$ epochs. It is a continuous metric with higher values indicating greater uncertainty for a training example.
	\begin{equation}
		f_\text{\textsc{var}}(x_i, y_i) = \sqrt{\frac{\sum_{e=1}^E \left( p_{\theta^{(e)}} \left(y_i | x_i \right) - \mu_i \right)^2}{E}}
		\label{eq:variability}
	\end{equation}

\end{itemize}

Confidence and correctness are the primary metrics that we use in our curricula since low and high values correspond to hard and easy examples respectively. On the other hand, variability is used as an auxiliary metric since only high scores clearly represent uncertain examples while low scores offer no important information on their own.

\subsection{Schedulers}
We consider both discrete and continuous schedulers.
Each scheduler is paired with the metric that is most suited, i.e. the discrete correctness metric combined with annealing and the continuous confidence metric is combined with competence.

\noindent The \textbf{\textsc{Annealing}} (\textsc{$\text{corr}_\text{anneal}$}) scheduler proposed by \citet{xu-etal-2020-curriculum}, assumes that training data are split into buckets $\{d_1 \subset D, \dots, d_K \subset D\} $ with possibly different sizes $|d_i|$.
In particular, we group examples into the same bucket if they have the same \textit{correctness} score (see Equation (\ref{eq:correctness})). In total, this results in $E+1$ buckets, which are sorted in order of increasing difficulty.
Training starts with the easiest bucket. We then move on to the next bucket by also randomly selecting $1/(E+1)$ examples from each previous bucket. Following prior work, we train on each bucket for one epoch.

\noindent The \textbf{\textsc{Competence}} (\textsc{$\text{conf}_\text{comp}$}) scheduler was originally proposed by \citet{platanios-etal-2019-competence}.
Here, we sort examples based on the \textit{confidence} metric (see Equation (\ref{eq:confidence})),
and use a monotonically increasing function to obtain the percentage of available training data at each step. The model can use only the top $K$ most confident examples as instructed by this function. A mini-batch is then sampled uniformly from the available examples.

In addition to those schedulers, we introduce the following modifications that take advantage of the \textit{variability} metric.
\textbf{\textsc{Correctness + Variability Annealing}} (\textsc{$\text{corr+var}_\text{anneal}$}) is a modification of the Annealing scheduler and \textbf{\textsc{Confidence + Variability Competence}} (\textsc{$\text{conf+var}_\text{comp}$}) is a modification of the Competence scheduler.
In both variations, instead of sampling uniformly across available examples, we give higher probability to instances with high \textit{variability} scores (Equation (\ref{eq:variability})), essentially using two metrics instead of one.
We assume that since the model is more uncertain about such examples further training on them can be beneficial.
For all curricula, after the model has finished the curriculum stage, we resume training as normal, i.e. by random sampling of training instances.


\subsection{Transfer Teacher Curriculum Learning}

In order to train a model (student) with training dynamics provided by another model (teacher), the latter should be first fine-tuned on a target dataset. In other words, the proposed metrics are used in a transfer teacher CL setting~\citep{matiisen2019teacher}.

\begin{figure}[t!]
	\centering
	\includegraphics[width=\linewidth]{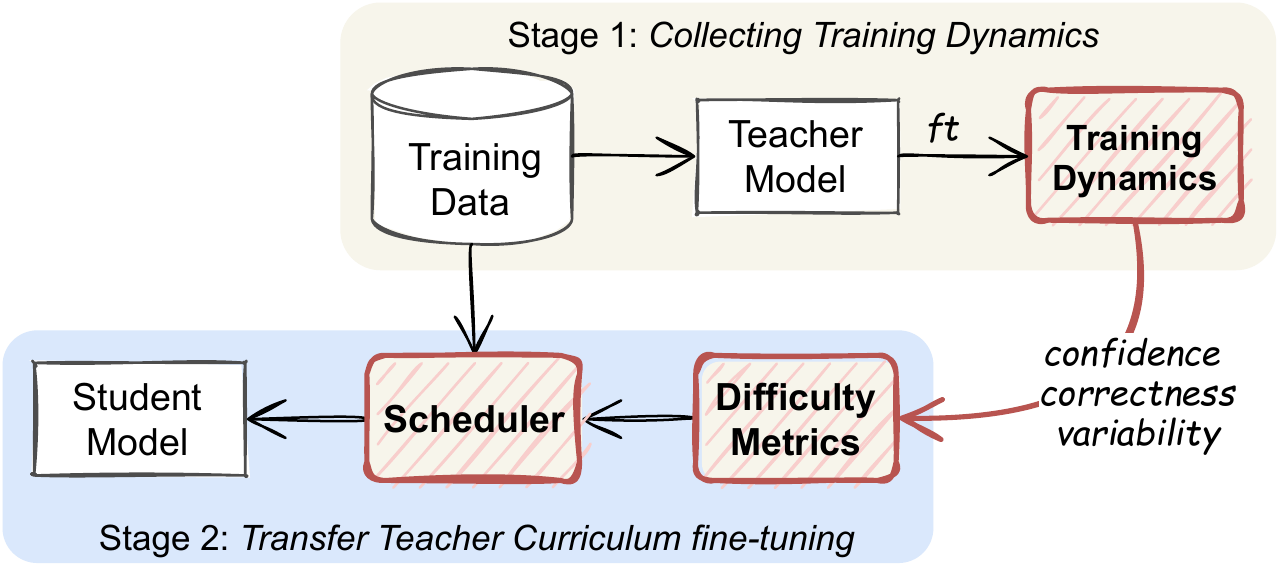}
	\caption{Transfer Teacher Curriculum Learning used in our study. A teacher model determines the difficulty of training examples by collecting training dynamics during fine-tuning (Stage 1). The collected dynamics are converted into difficulty metrics and are given to a student model via a scheduler (Stage 2).}
	\label{fig:curriculum}
\end{figure}

The two-step procedure that we follow in this study is depicted in Figure~\ref{fig:curriculum}. Initially a model (the \textit{teacher}) is fine-tuned on a target dataset and training dynamics are collected during the course of training. The collected dynamics are then converted into difficulty metrics, following Equations (\ref{eq:confidence})-(\ref{eq:variability}).
In the second stage, the difficulty metrics and the original training data are fed into a scheduler that re-orders the examples according to their difficulty (in our case from easy-to-hard) and feeds them into another model (the \textit{student}) that is the same in size and type as the teacher.

\section{Experimental Setup}

\begin{table*}[t!]
	\centering
	\scalebox{0.74}{
		\begin{tabular}{llllcrrrrr}
			\toprule
			\textsc{Train Set} & \textsc{ZS} & \sc{ID} & \textsc{OOD} & \textsc{\# Langs} & \textsc{\# Train} & \textsc{\# Val.} &  \textsc{\# ZS Test} & \textsc{\# ID Test} & \textsc{\# OOD Test} \\ \midrule

			PAWS & PAWS-X & PAWS & TwitterPPDB  &  7 &  49,401 & 2,000 & 2,000  & 8,000 & 9,324 \\
			MNLI & XNLI   & MNLI-m & NLI Diagnostics  &  15 &  392,702 & 2,490  & 5,010 & 9,815 & 1,104 \\
			SIQA  & XCOPA & SIQA & CSQA & 12 &  33,410 & 100  & 500 & 2,224  & 1,221 \\
			MLDoc &  MLDoc    &	-   &  -  &  8  & 10,000  & 1,000   & 4,000  & -  & - \\
			QNLI  &  - 	 &  QNLI   & Adversarial SQuAD  &  1  & 99,505    & 5,238   & -      & 5,463     &  7,857 \\
			RTE   &  -   & RTE     &  HANS  & 1  &  2,365  &  125   &  -     &  277      & 30,000   \\

			\bottomrule
		\end{tabular}
	}
	\caption{Datasets statistics. ZS, ID and OOD correspond to zero-shot Cross-lingual transfer, in-distribution and out-of-distribution settings, respectively. ZS Validation and Test statistics are per language.}
	\label{tab:data_stats}
\end{table*}

\subsection{Datasets}
\label{sec:datasets}

In this work we focus on four NLU classifications tasks: Natural Language Inference, where given a premise and a hypothesis the task is to identify if the hypothesis entails/contradicts/or is neutral based on the premise; Paraphrase Identification, where the task is to find if two sentences are paraphrases of one another; Commonsense Causal Reasoning, where given a premise, a question and a set of choices the task is to find the correct answer to the question based on the premise, and Document Classification where each document should be assigned the correct category.

We aim for a comparison across 3 settings: in-distribution (ID), out-of-distribution (OOD) and zero-shot (ZS), hence, we select datasets that contain all those settings, if possible.
We use a small subset from the GLUE benchmark~\citep{wang-etal-2018-glue} covering the NLI task (RTE, QNLI and MNLI) and four cross-lingual datasets: XNLI~\citep{conneau-etal-2018-xnli}, PAWS-X~\citep{yang-etal-2019-paws} for paraphrase detection, XCOPA~\citep{ponti-etal-2020-xcopa} for Commonsense reasoning and MLDoc~\citep{schwenk-li-2018-corpus} for document classification that combined cover 25 languages.
QQP and MRPC from GLUE were not included since PAWS is a paraphrase dataset with a cross-lingual version (PAWS-X).
The choice of the OOD datasets followed prior work: TwitterPPDB (as in~\citet{desai-durrett-2020-calibration}) for Paraphrase detection, NLI Diagnostics, Adversarial SQuAD and HANS for NLI (as in~\citet{swayamdipta-etal-2020-dataset}).
HANS was selected for RTE because both are binary classification datasets and there is no need to convert the ``neutral'' label to ``non-contradiction'' for evaluation.
CSQA was chosen as OOD for commonsense reasoning since it targets knowledge related to factual and physical commonsense, in contrast to SIQA or CosmosQA that focus on commonsense required during social/everyday situations.
Finally, for MLDoc we could not find a dataset having the same classification categories to serve as OOD. The corresponding statistics are shown in Table~\ref{tab:data_stats} and more details can be found in Appendix~\ref{app:datasets}.

\begin{table*}[h!]
\centering
	\scalebox{0.74}{
		\begin{tabular}{lcclccl}
			\toprule
			{\sc train} & \multicolumn{3}{c}{\sc paws}	& \multicolumn{3}{c}{\sc siqa}			\\
			 \cmidrule(lr){2-4} \cmidrule(lr){5-7}

			{\sc test}
			& {\sc paws} &  {\sc twitter}  &  \textit{Time} $\downarrow$
			& {\sc siqa}
			&  {\sc csqa}     &  \textit{Time} $\downarrow$
			\\
			& {\sc (id)} & {\sc (ood)} &
			& {\sc (id)} & {\sc (ood)} \\
			\cmidrule{1-1} \cmidrule(lr){2-4} \cmidrule(lr){5-7}

			\sc Random
			& 93.77~\textsubscript{0.30}
			& 72.18~\textsubscript{5.45} &
			& 68.36~\textsubscript{0.39}
			& 44.61~\textsubscript{0.96} &
			\\

			\sc $\text{CR}_\text{anneal}$
			& 93.72~\textsubscript{0.14}
			& 72.83~\textsubscript{6.65}  & 1.00
			& 68.45~\textsubscript{0.69}
			& 44.85~\textsubscript{0.72}   & 1.00
			\\

			\sc $\text{Corr}_\text{anneal}$
			& 93.93~\textsubscript{0.13}
			& 71.97~\textsubscript{2.69}	  & {0.56} ({0.35})
			& \textbf{69.20}~\textsubscript{0.48}
			& 45.81~\textsubscript{1.40} & 1.28	(1.11)
			\\

			\sc $\text{Conf}_\text{comp}$
			& 93.90~\textsubscript{0.11}
			& 75.18~\textsubscript{6.71}  & 1.28 (0.72)
			& 67.25~\textsubscript{1.80}
			& 43.93~\textsubscript{1.59}  & 1.13 (0.57)
			\\

			\sc $\text{Corr+Var}_\text{anneal}$
			& 93.82~\textsubscript{0.02}
			& 72.62~\textsubscript{1.17} & 0.77 (0.29)
			& 67.54~\textsubscript{0.43}
			& 44.31~\textsubscript{0.88} & {0.71}	({0.26})
			\\

			\sc $\text{Conf+Var}_\text{comp}$
			& \textbf{94.02}~\textsubscript{0.13}
			& \textbf{81.33}~\textsubscript{2.10}  & 1.20 (0.69)
			& 68.54~\textsubscript{0.04}
			& \textbf{45.84}~\textsubscript{0.67}  & 1.48	(0.71)
			\\

			\bottomrule
		\end{tabular}
	}

	\scalebox{0.74}{
		\begin{tabular}{lcclcclccl}
			\toprule
			{\sc train}
			& \multicolumn{3}{c}{\sc mnli}
			& \multicolumn{3}{c}{\sc rte}
			& \multicolumn{3}{c}{\sc qnli}
			\\
			\cmidrule(lr){2-4} \cmidrule(lr){5-7} \cmidrule(lr){8-10}

			{\sc test}
			& {\sc mnli-m} &  {\sc nli diag.}  &  \textit{Time} $\downarrow$
			& {\sc rte}    &  {\sc hans}       &  \textit{Time} $\downarrow$
			& {\sc qnli}   &  {\sc adv. squad} &  \textit{Time} $\downarrow$
			\\
			& {\sc (id)} & {\sc (ood)} &
			& {\sc (id)} & {\sc (ood)} &
			& {\sc (id)} & {\sc (ood)} & \\

			\cmidrule{1-1} \cmidrule(lr){2-4} \cmidrule(lr){5-7} \cmidrule(lr){8-10}

			\sc Random
			& 87.31~\textsubscript{0.22}
			& 61.47~\textsubscript{0.81} &
			& 75.57~\textsubscript{1.19}
			& 59.98~\textsubscript{2.66} &
			& 92.60~\textsubscript{0.18}
			& 81.07~\textsubscript{0.57} &
			\\

			\sc $\text{CR}_\text{anneal}$
			& 87.71~\textsubscript{0.16}
			& 61.56~\textsubscript{0.15} & 1.00
			& 74.01~\textsubscript{2.90}
			& 57.26~\textsubscript{3.18} & 1.00
			& 92.45~\textsubscript{0.27}
			& 81.94~\textsubscript{0.58} & 1.00
			\\

			\sc $\text{Corr}_\text{anneal}$
			& 87.53~\textsubscript{0.23}
			& 61.75~\textsubscript{0.59} & {0.76}	({0.47})
			& \textbf{76.17}~\textsubscript{1.06}
			& 55.15~\textsubscript{2.90} &  0.76 (0.57)
			& 92.57~\textsubscript{0.14}
			& 81.69~\textsubscript{0.82} & 1.30 (1.11)
			\\

			\sc $\text{Conf}_\text{comp}$
			& 87.36~\textsubscript{0.42}
			& 61.08~\textsubscript{0.72} & 1.33 (0.50)
			& 75.69~\textsubscript{1.62}
			& 55.05~\textsubscript{1.25} & 1.11 (0.78)
			& 92.68~\textsubscript{0.21}
			& 80.58~\textsubscript{0.51}	& 1.08 (0.89)
			\\

			\sc $\text{Corr+Var}_\text{anneal}$
			& 87.64~\textsubscript{0.03}
			& \textbf{62.05}~\textsubscript{0.41}   & 1.50	(0.81)
			& 75.45~\textsubscript{2.23}
			& 58.12~\textsubscript{5.76} & 1.00 (0.66)
			& \textbf{92.84}~\textsubscript{0.27}
			& \textbf{82.13}~\textsubscript{0.90} & 1.30 (1.00)
			\\

			\sc $\text{Conf+Var}_\text{comp}$
			& \textbf{87.74}~\textsubscript{0.27}
			& 61.56~\textsubscript{0.15} & 1.49 (0.60)
			& 76.05~\textsubscript{1.23}
			& \textbf{60.69}~\textsubscript{2.15} & 1.01 (0.78)
			& 92.63~\textsubscript{0.13}
			& 81.82~\textsubscript{0.34}	& 1.27 (1.07)
			\\

			\bottomrule
		\end{tabular}
	}
	\caption{Accuracy results of RoBERTa on in-distribution (ID) and out-of-distribution (OOD) data. \textit{Time} corresponds to the ratio $\nicefrac{S_{\text{*}_\text{TD}}}{S_{\text{CR}_\text{anneal}}}$, where the numerator is the number steps a curriculum with TD needs to reach the reported performance and the denominator is the number of steps the \textsc{$\text{CR}_\text{anneal}$} baseline requires to reach its performance. Results are reported over 3 random seeds and in parentheses we include the minimum time required across seeds.}
	\label{tab:OOD_mono}
\end{table*}


\subsection{Evaluation Settings}

We use the pre-trained versions of base RoBERTa~\citep{liu2019roberta} and XLM-R~\citep{conneau-etal-2020-unsupervised}.
For all datasets, we report accuracy as the main evaluation metric across three random seeds, on the following settings.

\noindent \textbf{\textsc{I}n-\textsc{D}istribution (\textsc{id})} and \textbf{\textsc{O}ut-\textsc{O}f-\textsc{D}istribution (\textsc{ood})}: We first fine-tune a monolingual (English) model on a target dataset and evaluate on their ID test set, e.g. train RoBERTa on MNLI, and evaluate on MNLI-M validation set. We also evaluate it on an OOD dataset, e.g. NLI Diagnostics.

\noindent \textbf{\textsc{Z}ero-\textsc{S}hot (\textsc{zs})}: Constitutes the zero-shot cross-lingual transfer setting. In particular, we train a multilingual model on the same dataset, e.g. XLM-RoBERTa on (English only) MNLI and evaluate it on a zero-shot cross-lingual set, e.g. XNLI test set~\citep{hu2020xtreme}.

In all experiments, we select the best checkpoint based on the \textit{English validation set} performance.
When reporting significance tests we use the Approximate Randomization test with all seeds~\citep{noreen1989computer}.
More details about experimental settings can be found in Appendix \ref{sec:curric_params}.

\subsection{Model Comparisons}

We primarily compare all curricula that use training dynamics against each other and against a baseline (\textit{Random}) that does not employ any curriculum and is using standard random order training.
We also consider as another baseline the teacher-transfer curriculum proposed by \citet{xu-etal-2020-curriculum}, namely \textit{Cross-Review} (indicated as \textsc{$\text{CR}_\text{anneal}$} in the next sections).
This curriculum uses the annealing scheduler, but does not employ training dynamics as difficulty scores.
Instead, the method splits the training set into subsets and a model is trained on each subset containing $1/N$ of the training set. The resulting models are then used to evaluate all examples belonging in different subsets. The difficulty score of an example is considered the number of its correct classifications across teachers.
We split each training set into $10$ subsets for all datasets except MLDoc where we split into $5$ and RTE where we split into $3$, following the original paper.
The difference between the \textit{cross-review} and the \textit{correctness} metrics is that Cross-Review uses $N$ fully trained teacher models on subsets of data, while the latter uses $E$ epochs of a single model trained on the entire training set.

Finally, when comparing \textsc{$\text{CR}_\text{anneal}$} with our training-dynamics based curricula, via discrete and continuous schedulers, we ensure that all of them are trained for equal amount of time, in order to have a one-to-one comparison. To enforce this, after the end of the curriculum phase, training continues as normal for the remaining steps (if any) by randomly sampling examples, otherwise training stops early.

\begin{table*}[t!]
	\centering
	\setlength{\tabcolsep}{5pt}
	\scalebox{0.75}{
		\begin{tabular}{@{}lllllllll@{}}
			\toprule
			{\sc train}
			& \multicolumn{1}{l}{\sc paws} &
			& \multicolumn{1}{l}{\sc mnli}  &
			& \multicolumn{1}{l}{\sc siqa}  &
			& \multicolumn{1}{l}{\sc mldoc}		 \\
			{\sc test}
			& \multicolumn{1}{l}{\sc paws-x (zs)} & \multicolumn{1}{c}{\textit{Time} $\downarrow$}
			& \multicolumn{1}{l}{\sc xnli (zs)} & \multicolumn{1}{c}{\textit{Time} $\downarrow$}
			& \multicolumn{1}{l}{\sc xcopa (zs)} & \multicolumn{1}{c}{\textit{Time} $\downarrow$}
			& \multicolumn{1}{l}{\sc mldoc (zs)} & \multicolumn{1}{c}{\textit{Time} $\downarrow$} \\
			\cmidrule{1-1}
			\cmidrule(lr){2-3} \cmidrule(lr){4-5}  \cmidrule(lr){6-7}  \cmidrule(lr){8-9}

			\sc Prior Work
			& 84.90$^*$ & -
			& 75.00$^*$ & -
			& 60.72 & -
			& 77.66 & - \\

			\sc Random
			& 84.49~\textsubscript{0.08}  &
			& 73.93~\textsubscript{0.18}  &
			& 60.62~\textsubscript{0.54}   &
			& \textbf{86.74}~\textsubscript{0.46} &  \\

			\sc $\text{CR}_\text{anneal}$
			&  84.35~\textsubscript{0.46} & 1.00
			&  74.57~\textsubscript{0.40} & 1.00
			&  60.44~\textsubscript{0.39} &	1.00
			&  86.59~\textsubscript{0.29} & 1.00		\\

			\sc $\text{Corr}_\text{anneal}$
			&  \textbf{84.70}~\textsubscript{0.15}  & 1.04 (0.85)
			&  73.92~\textsubscript{0.11}  & 1.11 (1.09)
			&  60.95~\textsubscript{0.40}  & 2.13 (0.77)
			&  86.47~\textsubscript{0.64} & 1.09 (1.02)  \\

			\sc $\text{Conf}_\text{comp}$
			&  84.51~\textsubscript{0.45}  & 1.44 (1.11)
			&  74.32~\textsubscript{0.41}   & 1.10 (0.53)
			&  61.09~\textsubscript{0.28} & 1.33 (0.8)
			&  86.30~\textsubscript{0.70} & 1.37 (1.18) \\

			\sc $\text{Corr+Var}_\text{anneal}$
			& 84.52~\textsubscript{0.27}   & {0.75} ({0.61})
			& \textbf{74.66}~\textsubscript{0.06}   & {0.79} ({0.49})
			&  \textbf{61.68}~\textsubscript{0.51}  & 2.73 (1.75)
			&  86.14~\textsubscript{0.23} &  {0.99} ({0.56})  \\

			\sc $\text{Conf+Var}_\text{comp}$
			&  84.03~\textsubscript{0.65}  & 1.50 (1.10)
			&  74.43~\textsubscript{0.18} & 1.17 (0.93)
			& 61.04~\textsubscript{0.31} & 1.32 ({0.58})
			&  85.78~\textsubscript{0.74} & 1.20 (0.94) \\

			\bottomrule
		\end{tabular}
	}
	\caption{Zero-shot performance between curricula as the average accuracy across languages (mean and standard deviation over 3 random seeds) with XLM-R.
		We also report prior work results for reference as follows: PAWS-X \citep{chi-etal-2022-xlm}, XNLI \citep{chi-etal-2022-xlm}, XCOPA \citep{ponti-etal-2020-xcopa}, MLDoc \citep{keung-etal-2020-dont} (mBERT).
		$^*$Note that \citet{chi-etal-2022-xlm} tune on the target languages validation sets.}
	\label{tab:zero_shot}
\end{table*}


\section{Experiments}

\begin{table*}[h!]
	\centering
	\scalebox{0.74}{
		\begin{tabular}{lccccccccc}
			\toprule
			\sc{train}
			& \multicolumn{3}{c}{\sc{paws}}
			& \multicolumn{3}{c}{\sc{mnli}}
			& \multicolumn{3}{c}{\sc{siqa}} \\
			\cmidrule(lr){2-4} \cmidrule(lr){5-7} \cmidrule(lr){8-10}

			\multirow{1}{*}{\sc{test}}
			& \sc{paws} & \sc{paws-x}  & \sc{twitter}
			& \sc{mnli-m} & \sc{xnli} & \sc{nli diag.}
			& \sc{siqa} & \sc{xcopa} & \sc{csqa}  \\

			& \sc{(id)} & \sc{(zs)} & \sc{(ood)}
			& \sc{(id)} & \sc{(zs)} & \sc{(ood)}
			& \sc{(id)} & \sc{(zs)} & \sc{(ood)}
			\\ \cmidrule(lr){1-1} \cmidrule(lr){2-4} \cmidrule(lr){5-7} \cmidrule(lr){8-10}

			\sc $\text{CR}_\text{anneal}$
			& 93.72~\textsubscript{0.14}
			& 84.35~\textsubscript{0.46}
			& 72.83~\textsubscript{6.65}
			& 87.71~\textsubscript{0.16}
			& 74.57~\textsubscript{0.40}
			& 61.56~\textsubscript{0.04}
			& 68.45~\textsubscript{0.69}
			& 60.44~\textsubscript{0.39}
			& 44.85~\textsubscript{0.72}
			\\

			\sc $\text{Corr}_\text{anneal}$
			& 93.93~\textsubscript{0.13}
			& \textbf{84.70}~\textsubscript{0.15}
			& 71.97~\textsubscript{2.69}
			& 87.53~\textsubscript{0.23}
			& 73.92~\textsubscript{0.11}
			& 61.75~\textsubscript{0.59}
			& \textbf{69.20}~\textsubscript{0.48}
			& 60.95~\textsubscript{0.40}
			& 45.81~\textsubscript{1.40}
			\\


			\sc $\text{Corr+Var}_\text{anneal}$
			& 93.82~\textsubscript{0.02}
			& 84.52~\textsubscript{0.27}
			& 72.62~\textsubscript{1.17}
			& 87.64~\textsubscript{0.03}
			& \textbf{74.66}~\textsubscript{0.06}
			& \textbf{62.05}~\textsubscript{0.41}
			& 67.54~\textsubscript{0.43}
			& \textbf{61.68}~\textsubscript{0.51}
			& 44.31~\textsubscript{0.88}
			\\

			\sc $\text{Conf+Var}_\text{comp}$
			& 94.02~\textsubscript{0.13}
			& 84.03~\textsubscript{0.65}
			& 81.33~\textsubscript{2.10}
			& \textbf{87.74}~\textsubscript{0.27}
			& 74.43~\textsubscript{0.18}
			& 61.56~\textsubscript{0.15}
			& 68.54~\textsubscript{0.04}
			& 61.04~\textsubscript{0.31}
			& \textbf{45.84}~\textsubscript{0.67}
			\\ \midrule

			\sc Length
			& 93.87~\textsubscript{0.31}
			& 84.56~\textsubscript{0.09}
			& 74.93~\textsubscript{5.66}
			& 87.22~\textsubscript{0.15}
			& 73.47~\textsubscript{0.29}
			& 61.20~\textsubscript{0.19}
			& 66.55~\textsubscript{1.45}
            & 60.76~\textsubscript{0.40}
            & 42.72~\textsubscript{0.72}
			\\

			\sc Rarity
			& \textbf{94.03}~\textsubscript{0.22}
			& 84.16~\textsubscript{0.24}
			& 79.90~\textsubscript{2.70}
			& 87.38~\textsubscript{0.10}
			& 73.42~\textsubscript{0.25}
			& 61.71~\textsubscript{0.56}
			& 66.29~\textsubscript{0.56}
			& 61.26~\textsubscript{0.43}
            & 42.67~\textsubscript{0.64}
			\\

			\sc PPL
			& 93.98~\textsubscript{0.23}
			& 84.09~\textsubscript{0.30}
			& \textbf{83.02}~\textsubscript{1.23}
			& 87.27~\textsubscript{0.10}
			& 73.42~\textsubscript{0.18}
			& 61.53~\textsubscript{0.67}
			& 68.27~\textsubscript{0.74}
			& 59.42~\textsubscript{1.18}
			& 44.69~\textsubscript{0.78}
			\\

			\bottomrule
		\end{tabular}
	}
	\caption{Task-specific (above the line) vs Task-agnostic metrics (below the line) on ID, ZS and OOD data.}
	\label{tab:metrics}
\end{table*}

\begin{figure*}[t!]
    \begin{subfigure}[b]{0.33\linewidth}
	    \includegraphics[width=\linewidth]{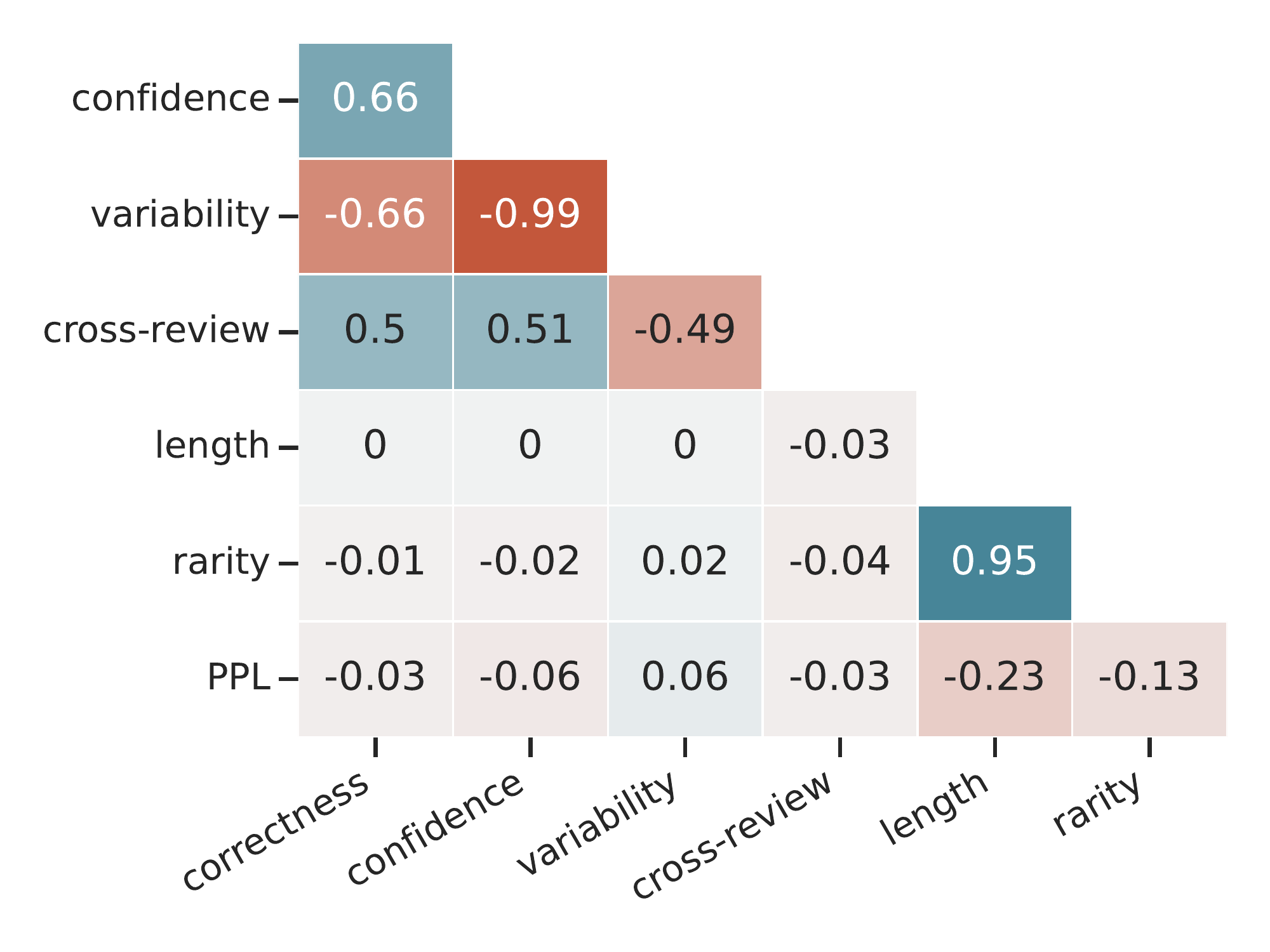}
	    \caption{PAWS}
	\end{subfigure}%
	\begin{subfigure}[b]{0.33\linewidth}
	    \includegraphics[width=\linewidth]{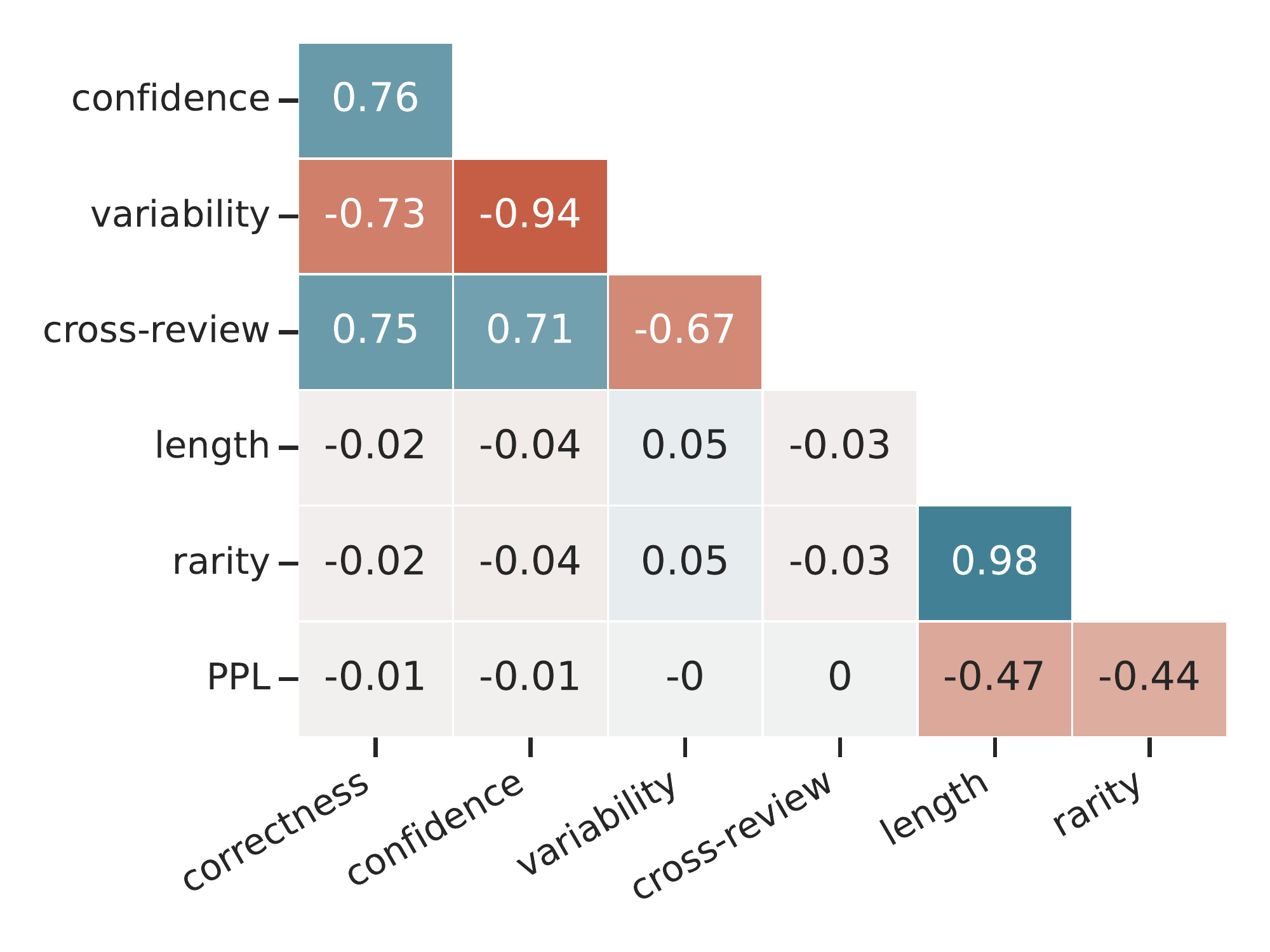}
	    \caption{MNLI}
	\end{subfigure}%
	\begin{subfigure}[b]{0.35\linewidth}
	    \includegraphics[width=\linewidth]{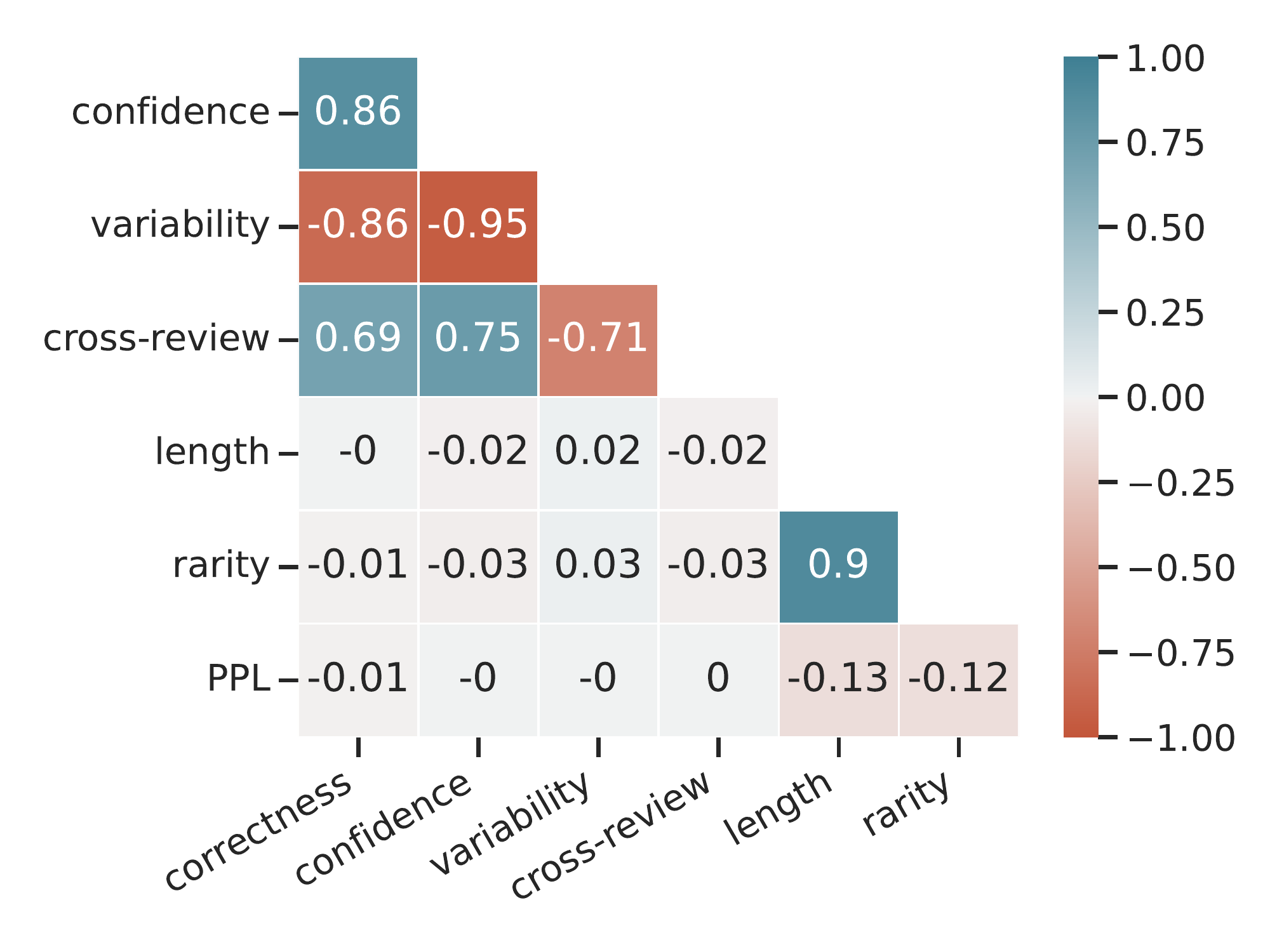}
	    \caption{SIQA}
	\end{subfigure}
	\caption{Spearman rank correlation between difficulty metrics using RoBERTa-base. Observations are similar for XLM-RoBERTa-base.}
	\label{fig:correlations_xlmr}
\end{figure*}

\subsection{Performance \& Training Time}
Results on Tables~\ref{tab:OOD_mono} and~\ref{tab:zero_shot} show performance and training time for various datasets.
The reported numbers (\textit{Time}) are calculated as the ratio $S_{\text{*}_\text{TD}}/S_{\text{CR}_\text{anneal}}$, i.e. the number of steps the Training Dynamics curriculum needs to reach best performance ($S_{\text{*}_\text{TD}}$) divided by the number of steps the Cross-Review method needs to reach its best performance ($S_{\text{CR}_\text{anneal}}$).
We focus comparison between curricula to show the trade-back between performance and time (a lower score indicates a larger speedup).
In parentheses the minimum time obtained across 3 random seeds is reported.

Table~\ref{tab:OOD_mono} shows accuracies for RoBERTa models when tested on ID/OOD data.
We observe that CL has minimal improvements in ID and in particular, through statistical testing we find that the increases over the Random baseline or Cross-Review are not significant for any of the datasets, except for MNLI-M vs Random \textsc{$\text{Conf+Var}_\text{comp}$}\footnote{After the time of the experiments, the SIQA test set was removed from public access. Thus, we also include dev set results in Appendix~\ref{sec:additional_results} for reproduction purposes.}.
Nevertheless, when tested on OOD performance improvement is larger.
\textsc{$\text{Conf+Var}_\text{comp}$} achieves the best performance on TwitterPPDB ($+9.15$, sign. $p<0.01$), CommonSenseQA ($+1.23$) and HANS ($+0.71$) while \textsc{$\text{Corr+Var}_\text{anneal}$} performs best for NLI Diagnostics ($+0.58$) and Adversarial SQuAD ($+1.06$, $p<0.01$) over random.
We speculate that \textsc{$\text{Conf+Var}_\text{comp}$} is better on OOD thanks to its slow pacing and the more accurate difficulties of confidence.
However, this comes at the cost of speedup by requiring either the same or a few more steps than \textsc{$\text{CR}_\text{anneal}$}.

Investigating the cross-lingual transfer results on Table \ref{tab:zero_shot}, initially we observe that CL with XLM-R seems to have a larger impact in terms of performance.
On XNLI there is a $+0.73$ points increase over Random ($p<0.01$). The difference with CR is not significant but TD achieved a 20\% speedup on average.
On XCOPA we observe $+1.06$ points increase, requiring however more training time with the \textsc{$\text{Corr+Var}_\text{anneal}$} curriculum, over the random baseline.
It is worth noting that for XCOPA, the competence-based curricula are able to also offer better performance with less additional training time.
As for the remaining datasets, CL is unable to achieve any performance improvement on MLDoc while on PAWS-X \textsc{$\text{Corr}_\text{anneal}$} has an improvement of $+0.2$ points from Random and $+0.35$ from \textsc{$\text{CR}_\text{anneal}$}, both statistically significant ($p<0.01$), with the cost of no speedup.
As another drawback, Cross-Review is generally more resource demanding since it needs $N$ fully-trained teacher models instead of $1$.

\subsection{Comparing Difficulties}

We now present a comparison between task-agnostic (TA) and task-specific (TS) difficulty metrics.
We re-implement 3 additional difficulty metrics proposed in prior work for Neural Machine Translation. The first two, introduced in \citet{platanios-etal-2019-competence}, correspond to sentence length (\textsc{length}) computed as the number of words in each sentence and word rarity (\textsc{rarity}) computed as the negated logarithmic sum of the frequency of each word in a sentence. Frequencies are computed over the training set.
Finally, we experiment with Perplexity (\textsc{ppl}) as the difficulty of a sentence~\citep{zhou-etal-2020-uncertainty}. We calculate sentence perplexity as the average perplexities of its subwords by masking one subword at a time and using the remaining context to predict it.
Since we test on a task with two-sentence input, we sum the \textsc{ppl} of the two sentences and consider the entire input for \textsc{length} and \textsc{rarity}.

Table \ref{tab:metrics} shows the results of the comparison between metrics on the PAWS and MNLI datasets.
Interestingly, we observe that TA metrics perform on par with TS on ID data, worse on ZS data and can perform quite well for OOD data. In particular, \textsc{rarity} is the third best on Twitter and the second best on NLI Diagnostics. This can be explained by the very different language used on Twitter vs Wikipedia in the training corpus, as well as the human-created nature of the NLI Diagnostics data.
\textsc{ppl} is the best performing system in Twitter and third best on CSQA. We find statistically significant improvement ($p<0.01$) compared with \textsc{$\text{Conf+Var}_\text{comp}$} on the Twitter OOD test set. Masked word prediction of unknown words could be an informative signal for a very new domain.
For the case of CSQA, length and rarity perform much worse than other metrics, possibly because the total length of the question and answer is quite small (approximately 15 tokens on average).

Furthermore, we analyse the relation of different difficulty metrics by calculating the Spearman rank correlation between all possible combinations.
As shown in Figure \ref{fig:correlations_xlmr}, we observe very high correlation between confidence and correctness, as expected, but also a good correlation with Cross-Review, explaining their close performance.
On the contrary, variability is negatively correlated with those metrics as higher values indicate more uncertainty from the model towards an example.
As such, a combination of these opposing metrics can offer benefits than combining two already correlated metrics.
Compared with task-agnostic metrics, interestingly, we see almost no (or negative) correlation with either \textsc{length}, \textsc{rarity} or \textsc{ppl}, indicating that examples that the model deems difficult when fine-tuned on a task are very different than those before fine-tuning or based on heuristics.
\textsc{rarity} and \textsc{length} highly correlate as longer sentences are more likely to contain rare words. Finally, \textsc{ppl} is reverse analogous to them, probably because longer sentences have more context and it is thus easier for the model to predict the masked token. Overall, \textsc{ppl} has a slight positive relation with variability since both measure model uncertainty and high \textsc{ppl} of words might make the model to further fluctuate between its predictions.

\subsection{Learning Curves}

\begin{figure}[t!]
	\centering
	\includegraphics[width=0.8\linewidth]{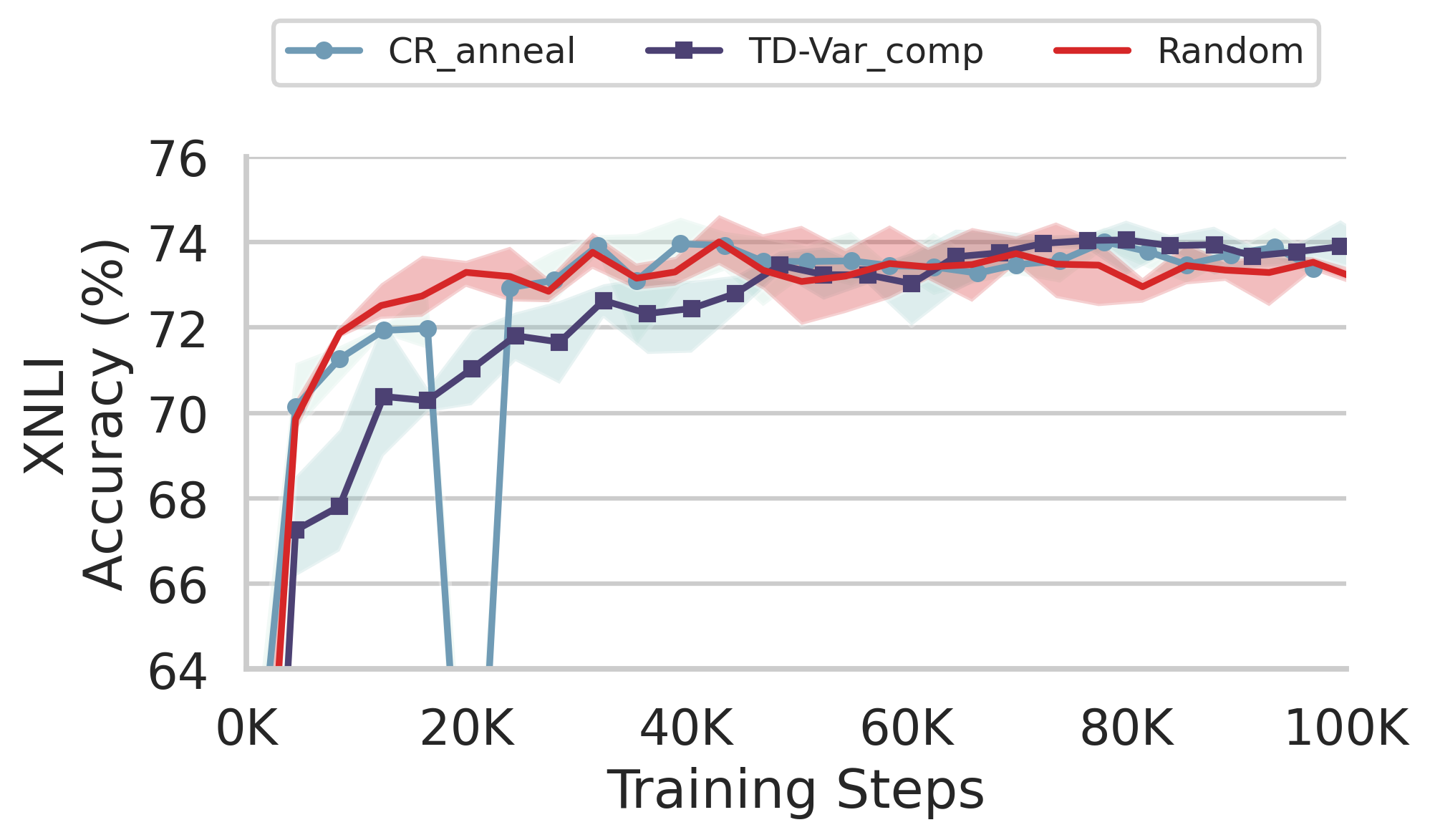}\\[-2pt]
	\includegraphics[width=0.8\linewidth]{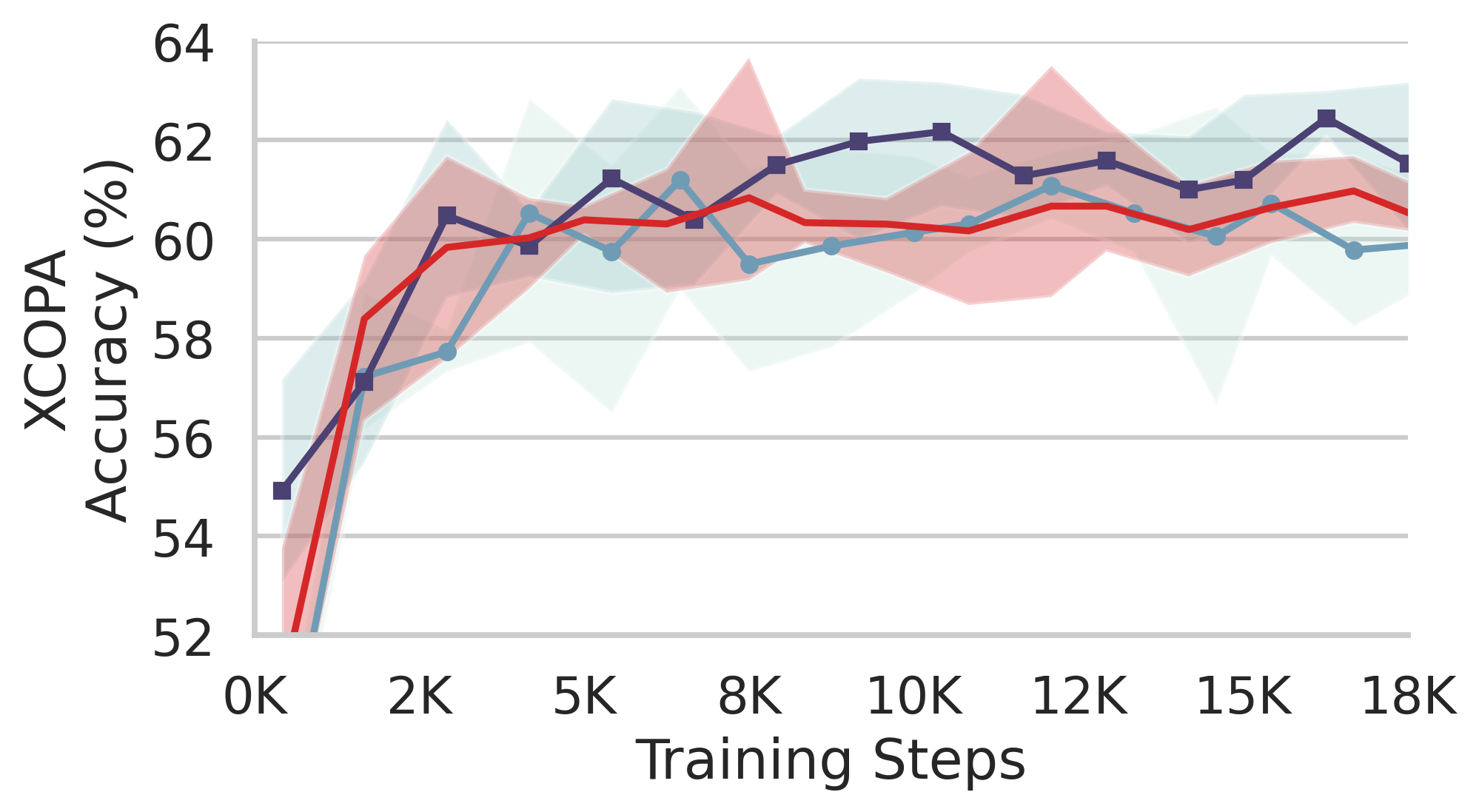}
	\caption{Average validation set accuracy across languages as a function of learning steps (in thousands) with XLM-R models. Results are reported over 3 random seeds.}
	\label{fig:learn_curves}
\end{figure}

In order to examine the behavior of the curricula during the course of training, we further plot the average language performance on the validation set as a function of the number of training steps when using XLM-R models for the improved datasets (XNLI and XCOPA).
In Figure \ref{fig:learn_curves} we draw the best performing curriculum (\textsc{$\text{Conf+Var}_\text{comp}$}), the \textsc{$\text{CR}_\text{anneal}$} and the Random baseline.

A first finding is that for \textsc{$\text{CR}_\text{anneal}$} we observe a performance drop around 20K steps in XNLI.
Further investigation revealed that the drop happens when the curriculum starts accessing the examples of the last bucket--which is the hardest one. This drop possibly indicates that buckets created by CR do not contain incrementally challenging examples that can help the model prepare for the hardest instances adequately, in contrast with training dynamics that result in smooth training.
In addition, we observe that after a point in training ($\approx$60K) random training stabilises while \textsc{$\text{Conf+Var}_\text{comp}$} continues to improve (70K-120K), despite having an initially lower performance than other schedulers.
Regarding XCOPA, the \textsc{$\text{Conf+Var}_\text{comp}$} curriculum is superior than random training and  \textsc{$\text{CR}_\text{anneal}$} by consistently improving performance from quite early in training (from step 8K onward).

\subsection{Training with limited budget}
\label{sec:budget}
Since training a teacher model can add overhead to the general training process (training a teacher model plus a similar-sized student), we further conduct a minimal experiment on PAWS, where we collect training dynamics for a teacher XLM-R model for different number of epochs (stopping training early) and then train a student  XLM-R model for longer, 10 epochs.
Results are reported in Table \ref{tab:minimal} for our best overall curriculum for this dataset \textsc{$\text{Corr+Var}_\text{anneal}$} as the average of the validation set languages performance.

We observe that it is not necessary to collect training dynamics for a long period of training (e.g. 10 epochs) as even with much less training, for instance 3 epochs, we can still get close performance to prior work much faster.
Compared to Cross-Review, that essentially requires full training of $N$ teacher models plus the student model, training dynamics offer a much more efficient solution.
Comparing training time with the \textsc{ppl} baseline, training dynamics are even faster as collecting sentence perplexities for the entire PAWS training set requires 1 hour and 30 minutes vs 36 minutes that are needed for 3 epochs of fine-tuning XLM-R.
Ultimately, even having less accurate dynamics (by training the teacher for less epochs) we can achieve overall less training time for the curriculum while still maintaining good performance.
Longer teacher training might be beneficial for future training of different student versions, e.g. a smaller-sized model.

\begin{table}[t!]
	\centering
	\scalebox{0.8}{
		\begin{tabular}{p{1.1cm}cccc}
			\toprule
			Teacher Epochs
			& \textsc{$\text{CR}_\text{anneal}$}
			& \textsc{$\text{Corr}_\text{anneal}$} & \textit{Time} $\downarrow$ \\ \midrule

			3 & \multirow{4}{*}{85.28~\textsubscript{0.18}}
			& 85.20~\textsubscript{0.17}  & 0.3 \\
			4 & & 85.46~\textsubscript{0.25} & 0.4 \\
			5 & & 84.94~\textsubscript{0.30} & 0.5 \\
			10 & & 85.34~\textsubscript{0.19}   & 1.0 \\

			\bottomrule
		\end{tabular}
	}
	\caption{Validation set performance (avg across languages) on PAWS-X with XLM-R models. Student is trained for 10 epochs, while training dynamics are collected from the teacher for different number of epochs.}
	\label{tab:minimal}
\end{table}

\section{Conclusion}

We presented a set of experiments using training dynamics~\citep{swayamdipta-etal-2020-dataset} as difficulty metrics for CL on several NLU tasks. Differently from existing works, we focus our evaluation on in-distribution, out-of-distribution and zero-shot cross-lingual transfer data by testing existing discrete and continuous schedulers as well as modifications of those in a transfer-teacher curriculum setting.

Our findings offer evidence that simply reordering the training examples in a meaningful way has mostly an impact on zero-shot cross-lingual transfer and OOD data, with no improvement on ID.
Our proposed Continuous scheduler with confidence and variability sampling provided a boost up to 8.5\% on a challenging OOD dataset over prior work.
Comparing our proposed application of training dynamics to other transfer-teacher curriculum methods that are using more than 1 teacher model, we observed greater speedups, improved performance and more stable training.
In particular, we found that task-agnostic metrics do not perform better than task-specific ones on ID and ZS data but can offer good performance on OOD settings.

Overall, our experiments suggest there is no curriculum outperforming others by a large margin which is consistent with findings in \citet{zhang2018empirical} and that task-agnostic metrics should not be rejected when transferring to challenging new domains. However we show that training dynamics are potentially better difficulty metrics for CL in both monolingual and multilingual models even with a limited budget.

Although in this study we focused on using CL on a ENglish only, a reasonable extension
is considering training data from other languages and investigate language-based instance difficulties or following efforts towards continual learning~\citep{parisi2019continual}.
Finally, using TD in a dynamic rather than a static curriculum is another interesting direction that can potentially offer further
training speedups as well as ways to improve model pre-training~\citep{nagatsuka-etal-2021-pre,li2021curriculum}.

\section*{Limitations}

The presented work has certain limitations that we acknowledge in this section. Firstly, the experiments are limited to base-sized models to enable us to conduct more experiments across multiple seeds. Validating that the same conclusions hold for large models is a promising direction.
The work is also focused on an offline curriculum approach, where difficulty metrics are obtained via the teacher model, before the student model training. This can indeed add an additional overhead to the overall process of collecting training dynamics. This limitation is partially addressed in Section~\ref{sec:budget}, to reduce overhead. Nevertheless, converting this approach into a dynamic one can be beneficial. Finally, following the original training dynamics setting, the methods were mainly applied on classification datasets, since it is straightforward to use accuracy as a difficulty metric.

\section*{Acknowledgements}
The authors thank the members of the Huawei Noah's Ark London NLP team for their comments on an early version of the paper.
We also thank the MindSpore team for providing technical support\footnote{\url{https://www.mindspore.cn/en}}\footnote{\url{https://github.com/mindspore-ai}}.

\bibliography{anthology,custom}
\bibliographystyle{acl_natbib}

\appendix

\section{Datasets}
\label{app:datasets}

In this study, we use the following datasets:
\begin{itemize}[leftmargin=-0.5pt, topsep=1pt, itemsep=0.1pt]
	\item[] \textbf{GLUE} \citep{wang-etal-2018-glue} is a benchmark for Natural language Understanding tasks. We use a subset of the included datasets: MNLI, RTE and QNLI that identify textual entailment (3 categories in the first one and 2 for the rest). Since the test set is hidden and results can be obtained only via submission to the benchmark, we sub-sample a 5\% portion from each training set and use it as our validation set. Then, final results are reported on the officially provided validation set.
	For the MNLI dataset, as ID test set we use the MNLI-matched validation set.
	As out-of-distribution test sets we use HANS~\citep{mccoy-etal-2019-right} for RTE, an evaluation set that tests if existing language models fail when using heuristics. We use Adversarial SQuAD~\citep{jia-liang-2017-adversarial} for QNLI, following \citet{swayamdipta-etal-2020-dataset}. Adversarial sentences that have high lexical overlap with the question, but do not answer it, have been inserted into the SQuAD~\citep{rajpurkar-etal-2016-squad} validation set. The original dataset follows the SQuAD format. We thus automatically convert it to a sentence-level one with binary labels, similarly to QNLI. Finally, we use NLI Diagnostics~\citep{wang-etal-2018-glue} as OOD test set for MNLI, a set of human-annotated examples that reveal model behavior on particular semantic phenomena.

	\item[] \textbf{PAWS-X} \citep{yang-etal-2019-paws} is the cross-lingual version of the English Paraphrase Adversaries from Word Scrambling dataset~\citep{zhang-etal-2019-paws} containing paraphrase identification pairs from Wikipedia. It consists of human translated pairs in 6 topologically distinct languages. The training set contains only English examples taken from the original PAWS dataset. As ID test set we use the test set of the original PAWS dataset. As OOD we use the TwitterPPDB dataset~\citep{lan-etal-2017-continuously} following \citet{desai-durrett-2020-calibration}.

	\item[] \textbf{XNLI} is the cross-lingual NLI dataset~\citep{conneau-etal-2018-xnli}, an evaluation set created by extending the development and test sets of the MultiNLI dataset~\citep{williams-etal-2018-broad} and translating it into 14 languages. Training data constitutes the original MultiNLI English training set.

	\item[] \textbf{XCOPA} is the Cross-lingual Choice of Plausible Alternatives~\citep{ponti-etal-2020-xcopa}, a typologically diverse multilingual dataset for causal common sense reasoning in 11 languages. The dataset consists of development and test examples for each language, which are translations from the English COPA~\citep{roemmele2011choice} validation and test sets.
	Following \citet{ponti-etal-2020-xcopa} we use the Social IQA dataset~\citep{sap-etal-2019-social} as training data (containing 3 possible choices), and the English COPA development set as validation data (containing 2 possible choices). For ID we report results on the SIQA test set (and validation set in Appendix~\ref{sec:additional_results}). For OOD, we consider the CommonSenseQA (CSQA) dataset~\citep{talmor-etal-2019-commonsenseqa} that contains 5 possible choices.

	\item[] \textbf{MLDoc} is a document classification dataset with 4 target categories: corporate/industrial, economics, government/social, and markets~\citep{schwenk-li-2018-corpus}. The dataset is an improved version of the Reuters benchmark~\citep{klementiev-etal-2012-inducing} consisting of 7 languages and comes with 4 different sets of English training data (1k, 2k, 5k, 10k). Here, we use the 10k following prior work~\citep{keung-etal-2020-dont}.

\end{itemize}

\section{Analysing Data Maps}

\begin{figure*}[t]
	\begin{subfigure}{0.25\linewidth}
		\includegraphics[width=\linewidth]{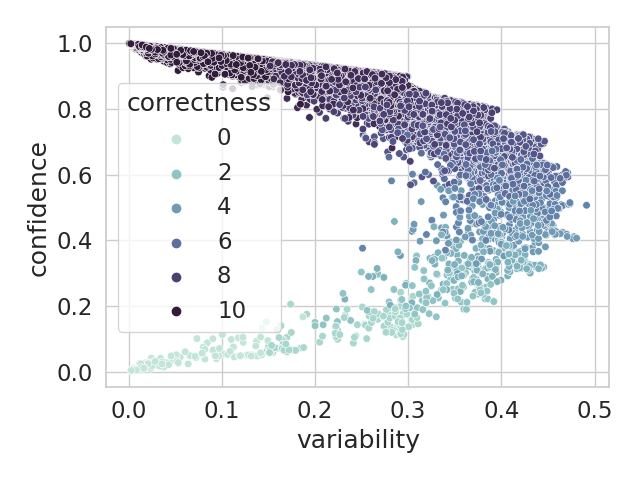}
		\caption{PAWS XLM-R}
		\label{fig:paws}
	\end{subfigure}%
    \begin{subfigure}{0.25\linewidth}
    	\includegraphics[width=\linewidth]{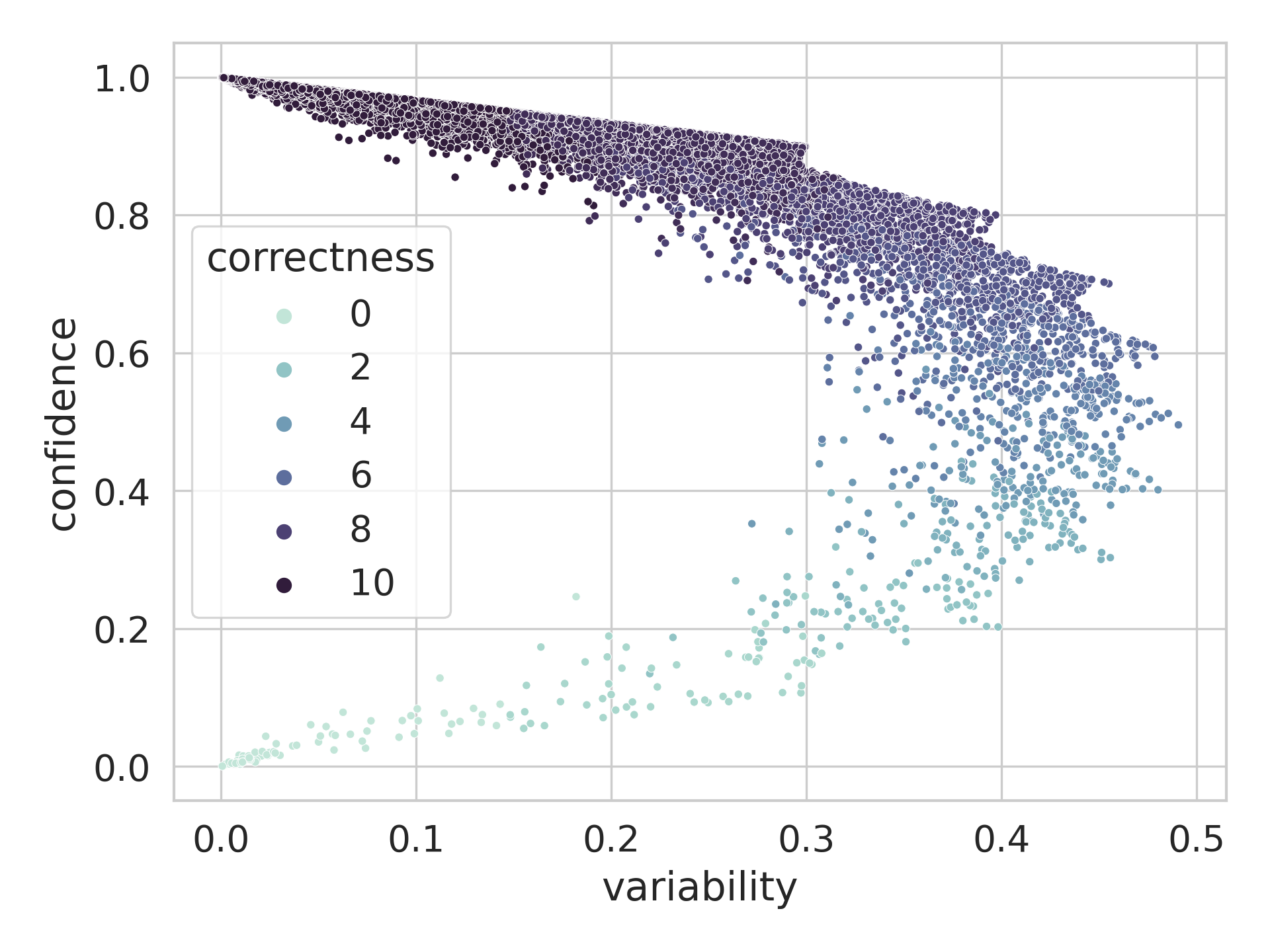}
    	\caption{PAWS RoBERTa}
    	\label{fig:paws}
    \end{subfigure}%
	\begin{subfigure}{0.25\linewidth}
		\includegraphics[width=\linewidth]{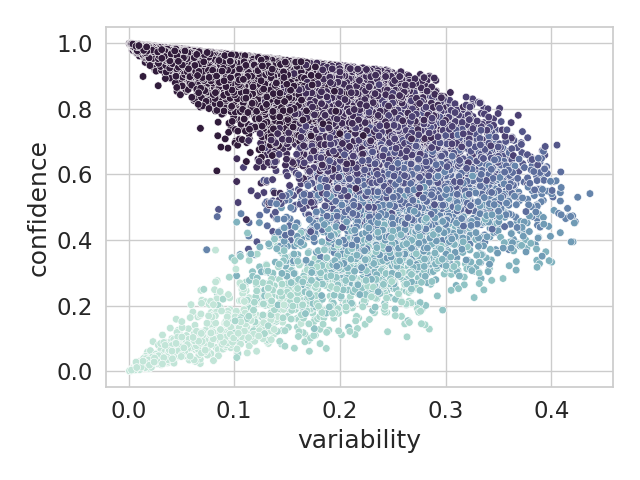}
		\caption{MNLI XLM-R}
		\label{fig:xnli}
	\end{subfigure}%
	\begin{subfigure}{0.25\linewidth}
		\includegraphics[width=\linewidth]{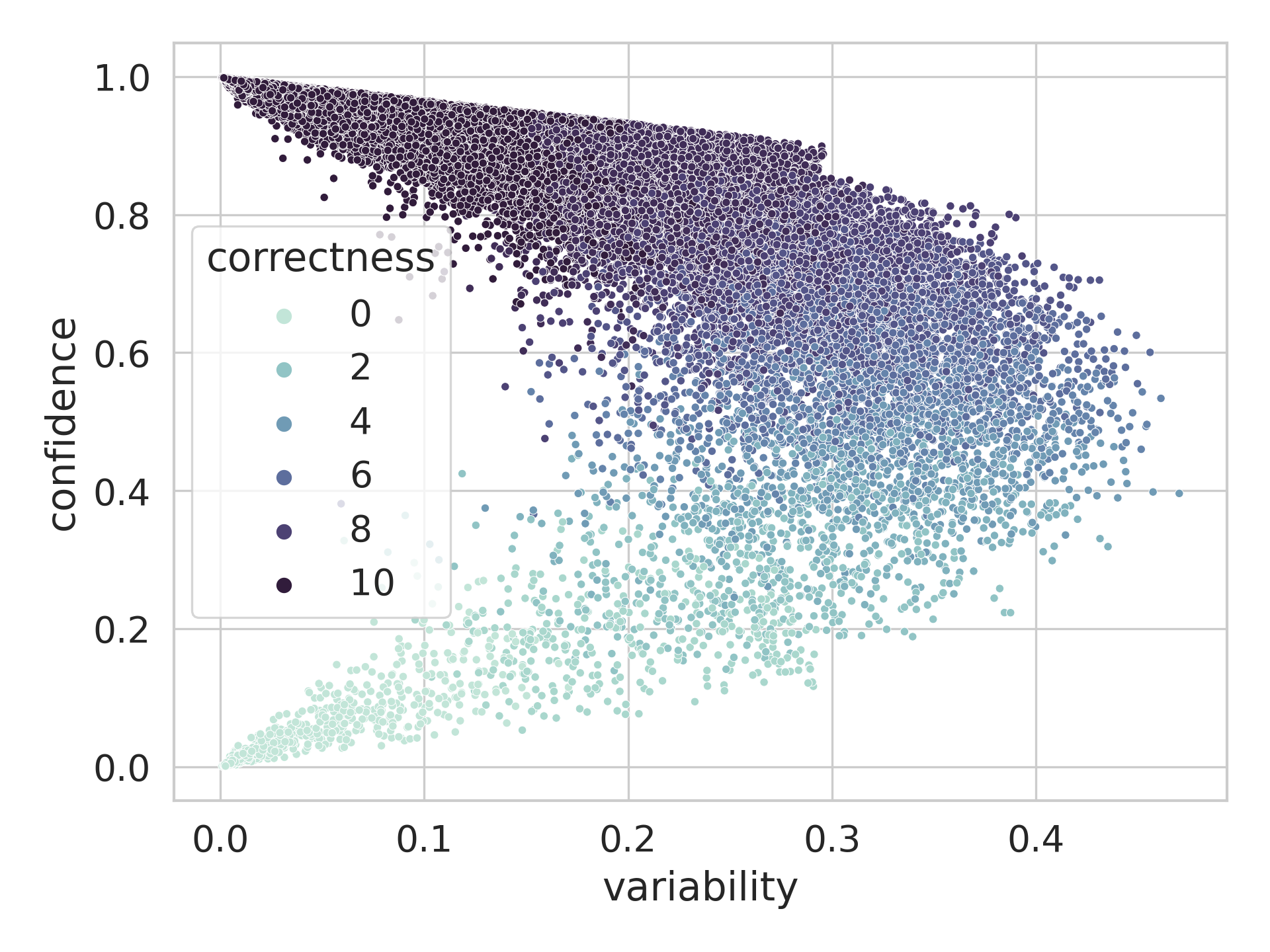}
		\caption{MNLI RoBERTa}
		\label{fig:xnli}
	\end{subfigure}
	\begin{subfigure}{0.25\linewidth}
		\includegraphics[width=\linewidth]{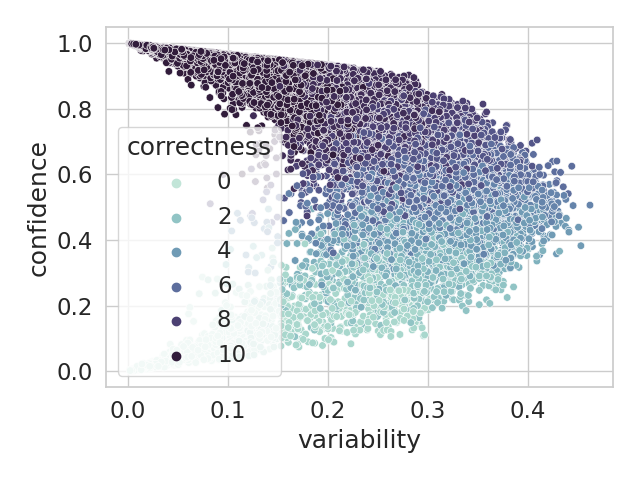}
		\caption{SIQA XLM-R}
		\label{fig:siqa}
	\end{subfigure}%
	\begin{subfigure}{0.25\linewidth}
		\includegraphics[width=\linewidth]{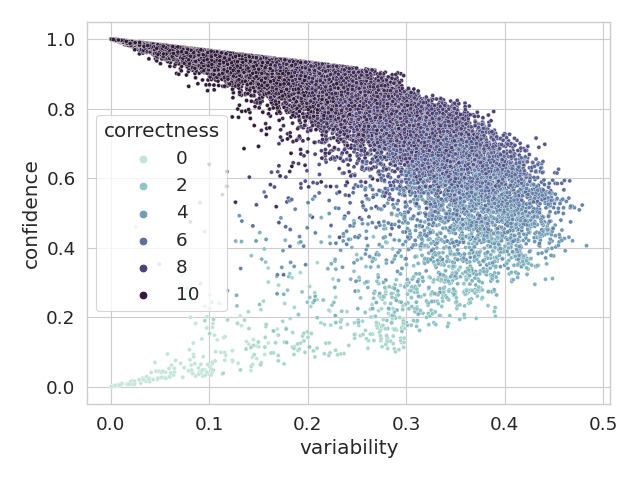}
		\caption{SIQA RoBERTa}
		\label{fig:siqa}
	\end{subfigure}

	\begin{subfigure}{0.25\linewidth}
		\includegraphics[width=\linewidth]{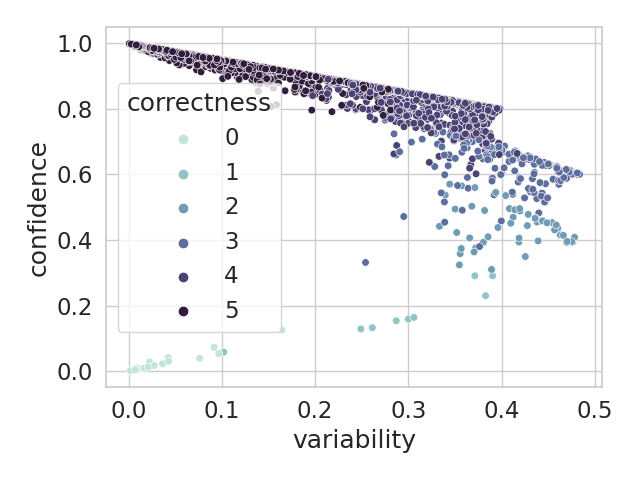}
		\caption{MLDoc XLM-R}
		\label{fig:mldoc}
	\end{subfigure}%
	\begin{subfigure}{0.25\linewidth}
	    \includegraphics[width=\linewidth]{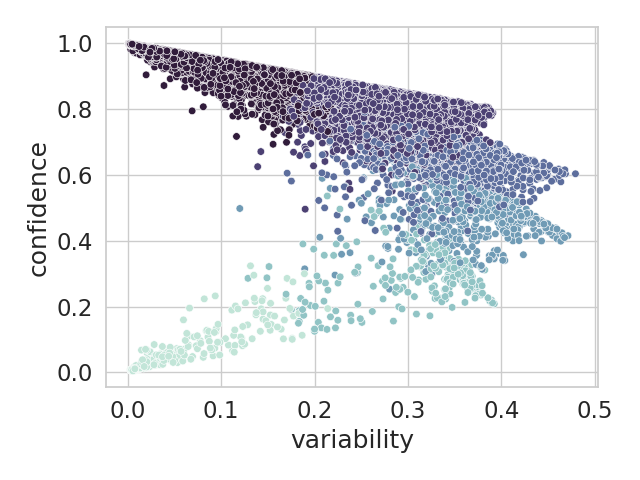}
	    \caption{QNLI RoBERTa}
	\label{fig:qnli}
    \end{subfigure}%
    \begin{subfigure}{0.25\linewidth}
    	\includegraphics[width=\linewidth]{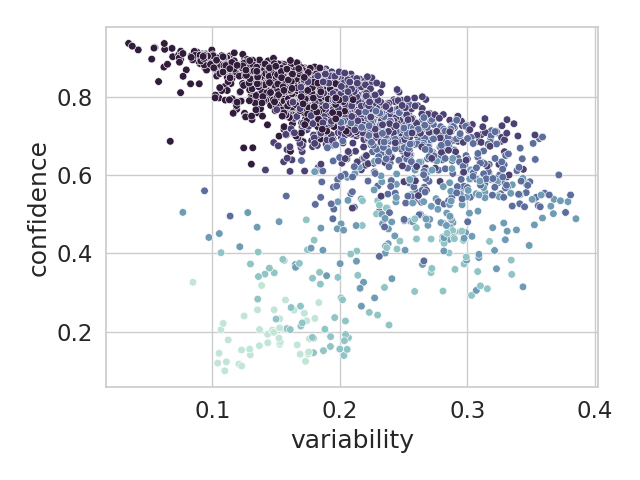}
    	\caption{RTE RoBERTa}
    	\label{fig:rte}
    \end{subfigure}
	\caption{Data map for the training set of each dataset. We plot maximum 25K examples for clarity.}
	\label{fig:data_maps}
\end{figure*}

To better understand the reason for the reported CL benefits we plot data maps that result from training an XLM-R model on each dataset in Figure \ref{fig:data_maps}, with confidence in the y-axis, variability in the x-axis and correctness in the legend.
As observed, the easiest overall datasets, i.e. PAWS-X  (\ref{fig:paws}), MLDoc (\ref{fig:mldoc}) and QNLI (\ref{fig:qnli}) result in quite crisp maps with very few hard-to-learn examples, while in XNLI (\ref{fig:xnli}) and SIQA (\ref{fig:siqa}) the data maps are very dense and the number of difficult examples is high. This can potentially explain why CL with XLM-R models was more beneficial on those datasets in terms of performance, confirming that CL can be used to better prepare a model for harder instances.

\section{Training Details}
\label{sec:training_details}

\subsection{Hyper-parameter Settings}
\label{sec:hyperparameters}
We use base models, XLM-R and RoBERTa with 470M and 340M parameters respectively, from the HuggingFace library~\citep{wolf-etal-2020-transformers}.
We fix sentence length to $128$ for all datasets except MLDoc where we use $256$.
We did minimal learning rate tuning on each dataset's English validation set, searching among  [7e-6, 1e-5, 2e-5, 3e-5] and choosing the best performing one, depicted in table \ref{tab:best_hypers}.
We clip gradients to $1.0$ after each update, use the AdamW optimizer~\citep{loshchilov2017decoupled} without any warmup.
All reported experiments use the same 3 random seeds and all models were trained on a single Nvidia V100 16GB GPU.
In terms of training time, Table \ref{tab:training_min} shows the training time required for each dataset with the above parameters.
\begin{table}
	\centering
	\scalebox{0.85}{
		\begin{tabular}{lrr}
			\toprule
			& $\text{RoBERTa}_{base}$ & $\text{XLM-R}_{base}$ \\ \cmidrule(lr){2-2} \cmidrule(lr){3-3}
			MNLI   &  7.5 h & 11.5 h   \\
			PAWS   &  1.0 h & 1.8 h\\
			SIQA   &  1.0 h & 1.3 h \\
			MLDoc  &  - & 1.0 h \\
			QNLI   &  1.0 h & - \\
			RTE    &  3 m & - \\
			\bottomrule
		\end{tabular}
	}
	\caption{Training time required for a full model training.}
	\label{tab:training_min}
\end{table}

\begin{table}[h!]
    \centering
    \scalebox{0.85}{
    \begin{tabular}{lccc}
        \toprule
        & Learning rate & Batch size & Epochs \\ \midrule
        MNLI & 7e-6 & 32 & 10 \\
        SIQA & 7e-6 & 8 & 10 \\
        PAWS-X & 1e-5 & 32 & 10 \\
        MLDoc & 3e-5 & 32 & 5 \\
        RTE & 2e-5 & 16 & 5 \\
        QNLI & 2e-5 & 32 & 5 \\
        \bottomrule
    \end{tabular}
    }
    \caption{Hyper-parameter settings for the used datasets}
    \label{tab:best_hypers}
\end{table}

\subsection{Multiple Choice QA}

We treat SIQA-XCOPA as a sentence-pair classification task and feed the model a (premise-question, choice) tuple converting each \textit{cause} into ``What was the cause?" and each \textit{effect} into ``What was the effect?" question which is concatenated to the premise.
Similar to prior work~\citep{ponti-etal-2020-xcopa} we use a feed forward linear layer on top of the input's first special token ($<$s$>$ in the case of RoBERTa and XLM-R) to produce a score for each of the possible choices.
In the case of CSQA that does not have a premise, we simply feed the network the question-choice pair.

\subsection{Curriculum Parameters}
\label{sec:curric_params}

In order to collect training dynamics we first fine-tune either a RoBERTa or an XLM-R model on the English training set of each dataset.
TD for each example are collected over the number of epochs that were employed for each dataset.
The \textsc{Competence} and \textsc{Competence Variability} schedulers require to set in advance the number of steps, i.e. total duration of the curriculum phase. We employ the same parameters as in \citet{platanios-etal-2019-competence} and set this value to $90\%$ of steps that the baseline model requires to achieve its best performance on the development set. The initial competence is set to $0.01$ for all datasets.
We evaluate each model at the end of each epoch and at regular intervals~\citep{dodge2020fine}, every 500 updates for MNLI (corresponding to 24 times per epoch) and 10 times per epoch for the rest of the datasets.
Performance is reported over three random seeds.

\section{Additional Results}
\label{sec:additional_results}

In Table~\ref{tab:siqa_more} we report test and validation set performance on the SIQA dataset.
We provide the per-language performance results for the multilingual datasets in Tables~\ref{tab:pawsx_per_lang}-\ref{tab:mldoc_per_lang}. Finally, we report per-category performance for the NLI Diagnostics dataset in Table~\ref{tab:nli_diag_more}.

\begin{table*}[h!]
\centering
\scalebox{0.85}{
    \begin{tabular}{lcc}
    \toprule
        & \sc SIQA dev set & \sc SIQA test set  \\ \midrule
        \sc Random
        &  69.70~\textsubscript{0.40}	       &  68.36~\textsubscript{0.39} \\

        \sc $\text{CR}_\text{anneal}$
	    &  68.76~\textsubscript{0.24}	   &  68.45~\textsubscript{0.69} \\

	    \sc $\text{Corr}_\text{anneal}$
	    & 69.28~\textsubscript{1.10}        & 69.20~\textsubscript{0.48} \\

	    \sc $\text{Conf}_\text{comp}$
	    & 68.73~\textsubscript{2.07}      & 67.25~\textsubscript{1.80} \\

	    \sc $\text{Corr+Var}_\text{anneal}$
	    & 69.24~\textsubscript{0.84}	   & 67.54~\textsubscript{0.43} \\

	    \sc $\text{Conf+Var}_\text{comp}$
	    & 69.21~\textsubscript{0.88}       & 68.54~\textsubscript{0.40} \\

	    \sc Length
	    &  68.51~\textsubscript{0.59}	   & 66.55~\textsubscript{1.45} \\

	    \sc Rarity
	    &  68.82~\textsubscript{0.36}	   & 66.29~\textsubscript{0.56} \\

	    \sc PPL
	    & 69.79~\textsubscript{0.74}      & 68.27~\textsubscript{0.74} \\
    \bottomrule
    \end{tabular}
    }
    \caption{SIQA development and test set results for different curricula. Results are averaged across 3 seeds.}
    \label{tab:siqa_more}
    \vspace{1.5cm}
\end{table*}

\begin{table*}[h!]
    \centering
    \scalebox{0.85}{
    \begin{tabular}{lcccccccc}
    \toprule
        & en	& 	fr	& 	es	& 	de	& 	zh	& 	ja		& ko		& AVG \\ \midrule

        \sc Random &
        94.58 & 	89.03 & 	88.80 & 	87.48 & 	80.08 & 	76.43 & 	75.03 & 	84.49 \\

        \sc $\text{CR}_\text{anneal}$  &
        94.57 & 	88.82 & 	88.5 & 	    87.03 & 	80.38 & 	76.20 & 	74.97 & 	84.35 \\

        \sc $\text{Corr}_\text{anneal}$  &
        94.73 & 	89.15 & 	88.73 & 	87.70 & 	81.02 &     76.43 & 	75.13 & 	84.70 \\

        \sc $\text{Corr}_\text{conf}$  &
        94.47 & 	88.62 & 	88.73 & 	87.13 & 	80.90 & 	76.18 &     75.52 & 	84.51 \\

        \sc $\text{Conf+Var}_\text{anneal}$  &
        94.53 & 	88.68 & 	88.45 & 	87.33 & 	80.65 & 	76.15 & 	75.83 & 	84.52 \\

        \sc $\text{Conf+Var}_\text{comp}$  &
        94.60 & 	88.4 & 	    88.32 & 	87.05 & 	80.23 & 	75.30 & 	74.30 & 	84.03 \\ \midrule

        \sc Length  &
        94.48 & 	88.65 & 	88.98 & 	87.15 & 	80.53 & 	76.07 & 	76.03 &     84.56 \\

        \sc Rarity  &
        94.38 & 	88.58 & 	88.47 & 	86.87 & 	80.20 & 	75.77 & 	74.87 & 	84.16 \\

        \sc PPL  &
        94.58 & 	88.75 & 	88.40 & 	87.40 &	    79.50 & 	75.92 & 	74.07 & 	84.09 \\

    \bottomrule
    \end{tabular}
    }
    \caption{Per language performance on the PAWS-X test set. Results are averaged across 3 seeds.}
    \label{tab:pawsx_per_lang}
    \vspace{1.5cm}
\end{table*}

\begin{table*}[h!]
    \centering
    \scalebox{0.65}{
    \begin{tabular}{lcccccccccccccccc}
    \toprule
        & en & fr & es & de & el & bg & ru & tr & ar & vi & th & zh & hi & sw & ur & AVG \\ \midrule

        \sc Random &
        84.60 & 	77.51 & 	78.36 & 	76.47 & 	75.62 & 	77.35 & 	75.36 & 	72.24 & 	72.01 & 	74.3 & 	71.64 & 	73.35 & 	69.62 &     64.86 & 	65.75 & 	73.93 \\

		\sc $\text{CR}_\text{anneal}$  &
        84.92 & 	77.96 & 	79.16 & 	77.21 & 	76.31 & 	77.71 & 	75.66	 & 73.19 & 	72.75 & 	75.00 & 	72.44 & 	74.15 & 	70.30 & 	65.15 & 	66.61 &     74.57  \\

		\sc $\text{Corr}_\text{anneal}$  &
        84.62 & 	77.74 & 	78.45 & 	76.74 & 	75.44 & 	77.28 & 	75.13 & 	72.36 & 	71.60 & 	74.03 & 	72.20 & 	73.41 & 	69.88 & 	64.32 & 	65.58 & 	73.92  \\

		\sc $\text{Corr}_\text{conf}$  &
        84.37 & 	77.65 & 	78.64 & 	76.57 & 	75.93 & 	78.07 & 	75.83	 & 73.18 & 	72.02 & 	74.78 & 	71.98 & 	73.51 & 	70.28 & 	65.93 & 	66.11 & 	74.32  \\

		\sc $\text{Conf+Var}_\text{anneal}$  &
        84.85 & 	78.02 & 	79.17 & 	77.29 & 	76.11 & 	78.05 & 	75.90 & 	72.95 & 	72.51 & 	75.06 & 	72.94 &     74.38 & 	70.50 &	    65.33 & 	66.86 &     74.66  \\

		\sc $\text{Conf+Var}_\text{comp}$  &
        85.07 & 	78.00 & 	78.62 & 	77.03 & 	76.44 & 	77.60 & 	76.11 & 	73.17 & 	72.22 & 	74.74 & 	72.32 & 	73.52 & 	69.95 & 	65.16 & 	66.56 & 	74.43  \\ \midrule

        \sc Length &
        84.26 &	77.84 &	78.13 &	75.93 &	74.91 &	77.31 &	74.54 &	71.78 &	71.11 &	74.11 &	71.47 &	72.73 &	69.29 &	64.07 &	64.54 &	73.47 \\

        \sc Rarity &
        84.38 & 	77.37 & 	78.07 & 	75.99 & 	75.00 & 	77.17 & 	74.46	 & 71.75 & 	71.12 & 	74.40 & 	71.47 & 	72.60 & 	68.86 & 	64.46 & 	64.20 & 	73.42  \\

        \sc PPL &
        83.84 & 	76.89	 & 78.00 & 	76.00 & 	74.69 & 	76.85 & 	75.01 & 	71.94 & 	71.74 & 	73.63 & 	71.26 & 	72.80 & 	69.03 & 	64.22 & 	65.35 & 	73.42  \\

    \bottomrule
    \end{tabular}
    }
    \caption{Per language performance on the XNLI test set. Results are averaged across 3 seeds.}
    \label{tab:xnli_per_lang}
    \vspace{1.5cm}
\end{table*}

\begin{table*}[h!]
    \centering\scalebox{0.8}{
    \begin{tabular}{lccccccccccccc}
    \toprule
        & en	& et	& 	ht & 	id & 	it	 & qu & 	sw & 	ta & 	th & 	tr & 	vi & 	zh & 	AVG \\ \midrule

        \sc Random &
        69.67 & 	56.47 & 	49.67 & 	66.93 & 	63.47	 & 49.93 & 	54.47 & 	64.00 & 	59.60 & 	60.20 & 	66.87 & 	66.20 & 	60.62 \\

        \sc $\text{CR}_\text{anneal}$  &
        68.80 & 	58.93 & 	51.27 & 	65.93 & 	63.73 & 	50.40 & 	53.13 & 	60.40 & 	60.07 & 	60.33 & 	65.33 & 	67.00 & 	60.44 \\

        \sc $\text{Corr}_\text{anneal}$  &
        67.67 & 	58.80 & 	50.47 & 	67.47 & 	63.20 & 	50.13 & 	56.00 & 	63.93 & 	60.60 & 	60.87 & 	65.33 & 	66.93 & 	60.95 \\

       \sc $\text{Corr}_\text{comp}$  &
        67.47 & 	58.00 & 	53.07 & 	66.20 & 	63.40 & 	52.20 & 	56.33 & 	63.20 & 	61.20 & 	61.87 & 	64.07 & 	66.13 & 	61.09 \\

        \sc $\text{Conf+Var}_\text{anneal}$  &
        69.33 & 	61.27 & 	50.47 & 	68.07 &     64.47 & 	49.80 & 	54.87 & 	64.40 & 	62.20 & 	60.93 & 	66.53 & 	67.87 & 	61.68 \\

    	\sc $\text{Conf+Var}_\text{comp}$  &
        66.93 &     58.60 & 	52.00 & 	66.40 & 	63.67 & 	52.07 & 	56.93 & 	61.00 & 	62.07 & 	59.47 & 	66.00 & 	67.33 & 	61.04 \\ \midrule

        \sc Length &
        67.33 & 	57.73 & 	50.20 & 	64.67 & 	64.07 & 	52.60 & 	55.00 & 	62.07 & 	63.93 & 	59.53 & 	65.20 & 	66.80 & 	60.76 \\

        \sc Rarity &
        68.73 & 	59.27 & 	51.27 & 	67.33 & 	62.67 & 	49.60 & 	55.73 & 	63.33 & 	62.60 & 	61.27 & 	65.80 & 	67.47 & 	61.26 \\

        \sc PPL &
        65.47 & 	56.53 & 	49.87 & 	63.53 & 	63.07 & 	50.87 & 	54.80 & 	61.13 & 	59.53 & 	60.27 & 	62.87 & 	65.07 & 	59.42 \\

    \bottomrule
    \end{tabular}
    }
    \caption{Per language performance on the XCOPA test set. Results are averaged across 3 seeds.}
    \label{tab:xcopa_per_lang}
\end{table*}

\begin{table*}[h!]
    \centering
    \scalebox{0.85}{
    \begin{tabular}{lccccccccc}
    \toprule
        & en & fr  &  de  &  es & 	it & 	ja & 	ru & 	zh & 	AVG \\ \midrule

        \sc Random &
        97.70 & 	92.41 & 	93.31 & 	87.45 & 	80.34 & 	80.72 & 	71.71 & 	90.32 & 	86.74 \\

        \sc $\text{CR}_\text{anneal}$  &
        97.57 & 	92.12 & 	93.23 & 	87.68 & 	80.40 & 	80.41 & 	71.76 & 	89.57 & 	86.59 \\

        \sc $\text{Corr}_\text{anneal}$  &
        97.57 & 	92.20 & 	92.93 & 	86.86 & 	80.57 & 	80.02 & 	71.84 & 	89.76 & 	86.47 \\

        \sc $\text{Corr}_\text{conf}$  &
        97.11 & 	91.37 & 	93.50 & 	87.33 & 	79.74 & 	80.34 & 	72.08 & 	88.90 & 	86.30 \\

        \sc $\text{Conf+Var}_\text{anneal}$  &
        97.52 & 	92.05 & 	93.03 & 	86.42 & 	80.61 & 	79.59 & 	70.74 & 	89.17 & 	86.14 \\

        \sc $\text{Conf+Var}_\text{conf}$  &
        97.09 & 	89.99 & 	93.50 & 	87.38 & 	79.86 & 	79.88 & 	70.88 & 	87.62 & 	85.78 \\

    \bottomrule
    \end{tabular}
    }
    \caption{Per language performance on the MLDoc test set. Results are averaged across 3 seeds.}
    \label{tab:mldoc_per_lang}
\end{table*}

\begin{table*}[t!]
    \centering
    \scalebox{0.9}{
    \begin{tabular}{lccccc}
    \toprule
        & Logic	& LS & PAS & Know & ALL \\ \midrule
        \sc Random
        & 55.31~\textsubscript{1.53}
        & 63.42~\textsubscript{0.56}
        & 66.98~\textsubscript{0.19}
        & 53.40~\textsubscript{1.09}
        & 61.47~\textsubscript{0.81}
        \\

        \sc $\text{CR}_\text{anneal}$
        & 55.59~\textsubscript{0.34}
        & 64.23~\textsubscript{0.84}
        & 67.45~\textsubscript{0.19}
        & 52.70~\textsubscript{0.33}
        & 61.56~\textsubscript{0.04}
        \\

		\sc $\text{Corr}_\text{anneal}$
		& 55.13~\textsubscript{0.56}
		& 63.96~\textsubscript{0.84}
		& 66.98~\textsubscript{0.51}
		& 54.81~\textsubscript{0.60}
		& 61.75~\textsubscript{0.59}
		\\

		\sc $\text{Conf}_\text{comp}$
		& 54.49~\textsubscript{1.35}
		& 63.33~\textsubscript{0.92}
		& 66.90~\textsubscript{0.40}
		& 53.05~\textsubscript{1.01}
		& 61.08~\textsubscript{0.72}
		\\

		\sc $\text{Corr+Var}_\text{anneal}$
		& 56.59~\textsubscript{1.03}
		& 64.23~\textsubscript{1.22}
		& 67.69~\textsubscript{0.39}
		& 52.35~\textsubscript{2.61}
		& 62.05~\textsubscript{0.41}
		\\

		\sc $\text{Conf+Var}_\text{comp}$
		& 55.68~\textsubscript{0.93}
		& 63.24~\textsubscript{0.44}
		& 67.53~\textsubscript{0.44}
		& 52.70~\textsubscript{0.44}
		& 61.56~\textsubscript{0.15}
		\\  \midrule

		\sc Length
		& 55.22~\textsubscript{0.39}
		& 64.50~\textsubscript{0.89}
		& 66.82~\textsubscript{0.68}
		& 51.29~\textsubscript{0.33}
		& 61.20~\textsubscript{0.19}
		\\

		\sc Rarity
		& 55.77~\textsubscript{0.22}
		& 64.68~\textsubscript{0.89}
		& 66.82~\textsubscript{0.40}
		& 52.00~\textsubscript{0.33}
		& 61.71~\textsubscript{0.56}
		\\

		\sc PPL
		& 56.50~\textsubscript{1.11}
		& 64.23~\textsubscript{0.67}
		& 67.53~\textsubscript{0.40}
		& 50.70~\textsubscript{0.76}
		& 61.53~\textsubscript{0.67}
		\\

	\bottomrule
    \end{tabular}
    }
    \caption{NLI Diagnostics results. Results are averaged across 3 seeds.}
    \label{tab:nli_diag_more}
\end{table*}

\end{document}